\documentclass{article} 
\usepackage{slim_arxiv,times}

\usepackage{amsmath,amsfonts,bm}

\def\eqref#1{equation~\ref{#1}}

\def\1{\bm{1}}

\DeclareMathAlphabet{\mathsfit}{\encodingdefault}{\sfdefault}{m}{sl}
\SetMathAlphabet{\mathsfit}{bold}{\encodingdefault}{\sfdefault}{bx}{n}

\usepackage{graphicx}
\usepackage{wrapfig}
\usepackage{colortbl}
\usepackage{amssymb}
\usepackage{hyperref}
\usepackage{url}
\hypersetup{hidelinks}

\definecolor{rankone}{HTML}{9CC9EA}
\definecolor{ranktwo}{HTML}{CBE3F5}
\definecolor{rankthree}{HTML}{EEF7FD}
\definecolor{dropred}{HTML}{C62828}
\newcommand{\best}[1]{\cellcolor{rankone}\textbf{#1}}
\newcommand{\second}[1]{\cellcolor{ranktwo}\textbf{#1}}
\newcommand{\third}[1]{\cellcolor{rankthree}#1}

\newcommand{\gapthickhline}{\noalign{\vskip 1pt\hrule height 0.8pt\vskip 1pt}}
\newcommand{\gaphline}{\noalign{\vskip 1pt\hrule height 0.4pt\vskip 1pt}}
\newcommand{\cmark}{\ensuremath{\checkmark}}
\newcommand{\xmark}{\ensuremath{\times}}

\title{SLIM-0.5B: Learning Action-Grounded Predictive Latents for Robot Manipulation}

\author{
\begin{minipage}{0.96\textwidth}
\centering
\textbf{Jingkai Wang}$^{1,2,*}$ \quad
\textbf{Zihan Tang}$^{3,2,*}$ \quad
\textbf{Gu Zhang}$^{3}$ \quad
\textbf{Mingyu Cao}$^{2}$
\\[0.25em]
\textbf{Jiapeng Chen}$^{1}$ \quad
\textbf{Jingjiao Zhao}$^{4,2}$ \quad
\textbf{Xiansheng Chen}$^{2}$ \quad
\textbf{Pengwei Wang}$^{2}$
\\[0.25em]
\textbf{Lemao Liu}$^{1}$ \quad
\textbf{Dejing Dou}$^{1}$
\\[0.65em]
\normalfont\small
$^{1}$Fudan University \quad
$^{2}$Beijing Academy of Artificial Intelligence
\\
$^{3}$Tsinghua University \quad
$^{4}$Renmin University of China
\\[0.4em]
$^{*}$Equal contribution.
\\[0.25em]
\href{https://kzz1031.github.io/slim-project-page/}{\textcolor{blue}{\texttt{https://kzz1031.github.io/slim-project-page/}}}
\end{minipage}
}

\begin{document}

\maketitle
\fancyhead{}

\begin{center}
    \includegraphics[width=\linewidth]{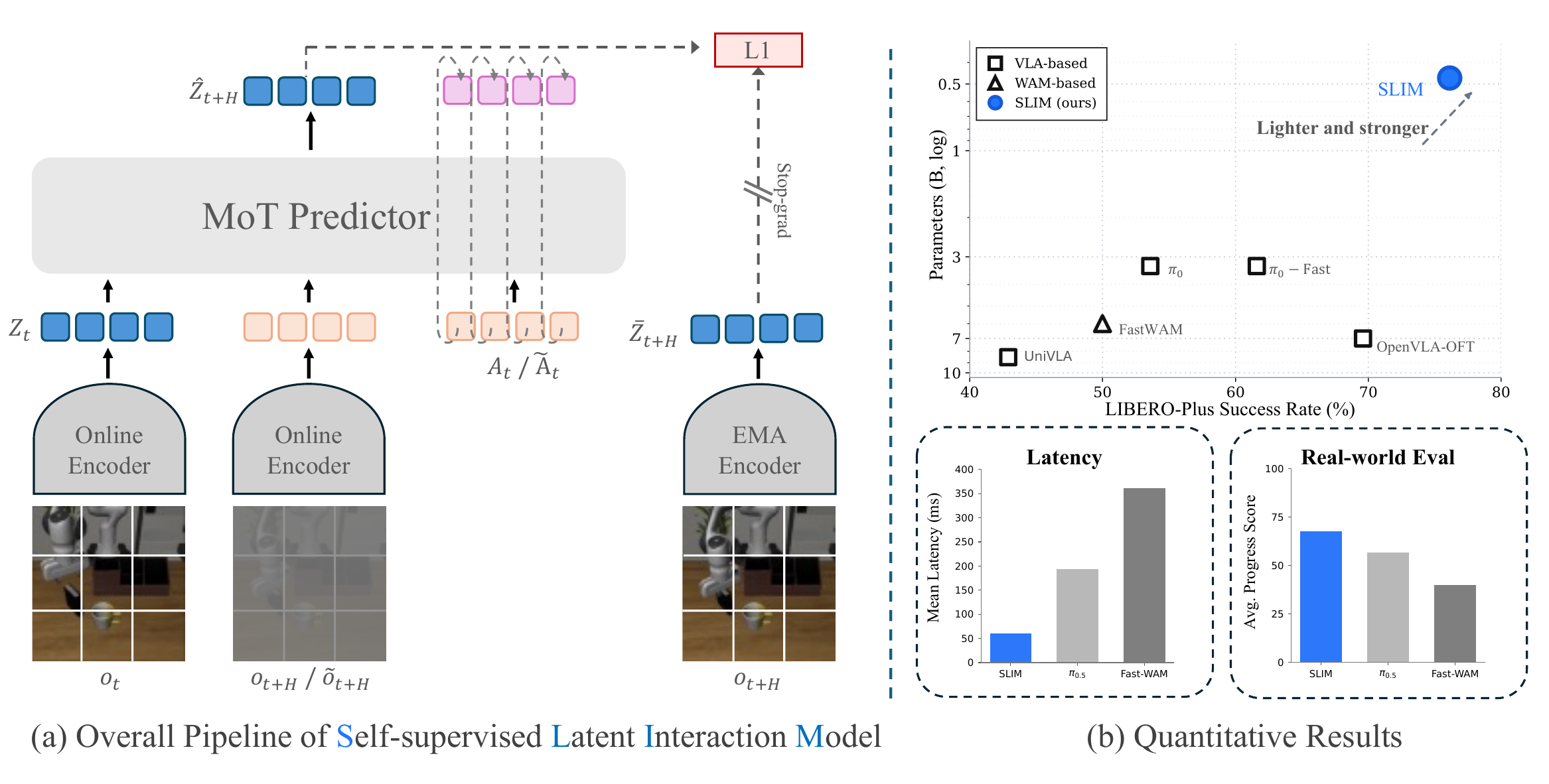}
    \refstepcounter{figure}
    \label{fig:teaser}
    \vspace{0.5em}
    \parbox{0.92\linewidth}{\small\centering
        \textbf{Figure~\thefigure: Overview of SLIM.}
        A compact MoT backbone models interactions among observation latents, continuous action tokens, and predictive future slots.
        Complementary inverse- and forward-dynamics objectives learn action-grounded predictive latents from robot trajectories, enabling robust, low-latency control with a compact policy.
    }
\end{center}

\begin{abstract}
Vision-language-action policies rely on large multimodal backbones to jointly perform perception, language conditioning, and action generation at every control step.
Much of this capacity supports open-domain semantics, whereas continuous robot manipulation primarily requires compact representations of observations, actions, and the transitions induced by actions. 
Pixel-level world models provide another route, but predicting visual details irrelevant to control can be unnecessarily expensive.
We propose \textbf{SLIM} (\textbf{S}elf-supervised \textbf{L}atent \textbf{I}nteraction \textbf{M}odel), a compact 0.5B-parameter latent interaction policy.
SLIM learns action-grounded predictive latents that capture both action-conditioned future transitions and the actions that explain observed changes.
SLIM learns these representations through self-supervised masked trajectory prediction, combining action reconstruction with future-latent prediction.
A compact Mixture-of-Transformers (MoT) backbone models interactions between observation latents and action tokens. The resulting policy is trained with flow matching for language-conditioned action generation.
Across simulation benchmarks and real-world evaluation, SLIM matches or exceeds representative large-scale VLA and world-action-model baselines with fewer parameters, no additional embodied pretraining, lower inference latency, and substantially lower GPU memory usage.
\end{abstract}

\section{Introduction}
\label{sec:introduction}

Language-conditioned robot manipulation is increasingly shaped by two modeling
paradigms. The first treats robot control as an extension of large
vision-language models: vision-language-action (VLA) policies place perception,
instruction understanding, and action generation within a shared multimodal
backbone
\citep{brohan2023rt2,driess2023palme,kim2024openvla,black2024pi0,pi05_2025,grootn1_2025}.
This strategy brings strong semantic priors and open-domain language
understanding to robot manipulation. However, the relation between an action
and the observation change it induces is typically learned only implicitly
through action supervision, rather than explicitly encoded in the policy
representation.

The second paradigm introduces predictive structure through world models and world-action models. 
These methods predict future images, videos, or joint video-action trajectories conditioned on actions
\citep{wu2023gr1,unifiedwm2025,ye2026dreamzero}. 
This direction is well aligned with robot control: actions should be grounded in the outcomes they cause.
However, pixel- or video-level prediction requires modeling appearance details that may be irrelevant to the next action decision, such as background textures, lighting, and minor visual changes, thus decreasing the generalization ability. Moreover, when future generation or world-model rollout is performed during inference, it introduces additional computation and latency into the control loop.


Together, these limitations motivate a different design point: a compact policy representation that retains semantic task conditioning while explicitly modeling action--observation interactions.
We call such a representation \emph{action-grounded}: it should support both inferring which action explains an observed transition and predicting the latent transition induced by an action.
Future prediction is therefore used as a training signal to shape the policy representation in latent space, rather than as the output of a pixel-generative model.

We instantiate this design as \textbf{SLIM}, a \textbf{S}elf-supervised \textbf{L}atent \textbf{I}nteraction \textbf{M}odel.
SLIM separates semantic conditioning from control computation: language instructions specify the task, observation latents represent the current state, and a compact Mixture-of-Transformers (MoT) backbone \citep{liang2025mot} models interactions between observation latents and continuous action tokens.
By learning action-conditioned transitions directly in the observation-latent space consumed by the policy, SLIM makes dynamics an explicit training objective without introducing a pixel decoder or a separate world-model state space.

SLIM realizes this bidirectional constraint through self-supervised masked trajectory prediction.
In the action-masked branch, the clean action chunk is replaced by its
flow-noised counterpart and reconstructed from the current and future
observation latents, yielding an inverse-dynamics objective.
In the future-latent-masked branch, learned mask embeddings replace the
future observation latent, which is then predicted from the current
observation latent and the clean action chunk, yielding an
action-conditioned forward-dynamics objective.
The two objectives ground the representation in both directions: \textbf{actions must explain observation changes, while future observation latents must remain predictable from actions.}
SLIM then reuses this bidirectional predictive structure to train a
compact flow-matching policy \citep{chi2023diffusionpolicy,liu2024rdt,black2024pi0}.

We evaluate SLIM across broad simulation benchmarks and real-world manipulation experiments \citep{liu2023libero,fei2025liberoplus,mees2021calvin}.
Across these settings, SLIM is competitive with or stronger than representative VLA and world-action-model baselines with substantially fewer parameters, lower inference latency, and substantially lower GPU memory usage.
The results suggest that robot manipulation does not need the main control computation to be either open-domain VLM reasoning or pixel-level future generation.
A compact action-grounded latent interaction process can provide a more efficient control backbone.

Our contributions are threefold.
\textbf{First}, we propose SLIM, a compact 0.5B-parameter latent interaction policy that places the main control computation in action-observation latent interaction.
\textbf{Second}, we introduce action-grounded masked trajectory prediction, which learns predictive observation latents through complementary action-chunk reconstruction and future-latent prediction objectives.
\textbf{Third}, we demonstrate strong performance on LIBERO, zero-shot LIBERO-Plus, CALVIN ABC$\rightarrow$D, and real-world manipulation, while using fewer parameters and achieving lower inference latency than representative VLA and world-action-model baselines.

\section{Related Work}
\label{sec:related-work}

\textbf{Vision-language-action policies.}
Vision-language-action (VLA) and generalist robot policies combine visual perception, language understanding, and action generation in a shared model.
Large-scale robot transformers established this paradigm across diverse demonstrations and embodiments \citep{brohan2022rt1,brohan2023rt2,driess2023palme,openx2023rtx}, followed by open-source and increasingly general VLA systems \citep{octo2024,kim2024openvla,black2024pi0,pi05_2025,grootn1_2025}.
Although VLAs benefit from broad visual-language knowledge, they typically retain a large VLM-style backbone at the center of every control step.
Continuous robot actions are commonly modeled with diffusion or flow matching \citep{chi2023diffusionpolicy,liu2024rdt}; recent VLAs adopt the same action-generation recipe while keeping the multimodal backbone for control.
SLIM also adopts flow matching for action generation, but places the main control computation in a compact backbone that models interactions between observation latents and action tokens.

\textbf{World action models and future-oriented policy representations.}
World action models connect actions with their consequences by predicting future observations or joint video-action trajectories \citep{wu2023gr1,unifiedwm2025,ye2026dreamzero}.
Their motivation is well aligned with robot control, but pixel-level future prediction can spend capacity on appearance details that are not needed for the next decision, and explicit future imagination at inference time can add control-loop latency.
Recent work reduces this cost by removing test-time video prediction, compressing visual bandwidth, or replacing pixel futures with latent subgoals \citep{fastwam2026,onetokenperframe2026,lawam2026}.
Closest to our setting, latent WAMs argue that compact latent targets are more useful for control than reconstruction-oriented futures \citep{lawam2026,reconstruction_semantics2026, repwam2026}.
A parallel line instead supervises policy representations with future targets from frozen encoders or perception teachers \citep{lastzero2026,lda1b2026,dreamvla2025}.
SLIM differs in two respects: it does not treat control as a WAM-conditioned policy or as alignment to a fixed external representation. Instead, it learns an action-grounded predictive latent through bidirectional masked trajectory prediction inside the same compact MoT used for control, and the deployed policy does not require generating future frames or conditioning on an explicit future latent.

\textbf{JEPA and action-grounded predictive representations.}
Joint-embedding predictive architectures (JEPA) learn representations by predicting masked targets in embedding space rather than reconstructing pixels \citep{lecun2022ami,assran2023ijepa}, and video extensions apply this principle to temporal understanding and prediction \citep{bardes2024vjepa,vjepa2_2025,vjepa21_2026}.
This view is closely related to actionable representation learning, which seeks to preserve factors relevant to decisions and control \citep{ghosh2018actionable}.
In robotics, VLA-JEPA adds JEPA-style future-latent prediction on a large VLM
backbone \citep{vlajepa2026}, while DeFI learns forward and inverse dynamics with separate modules \citep{disentangled_dynamics2026}.
SLIM instead couples both directions in one compact MoT and keeps prediction
in the observation-latent space used for control.

\begin{figure}[t]
    \centering
    \includegraphics[width=\linewidth]{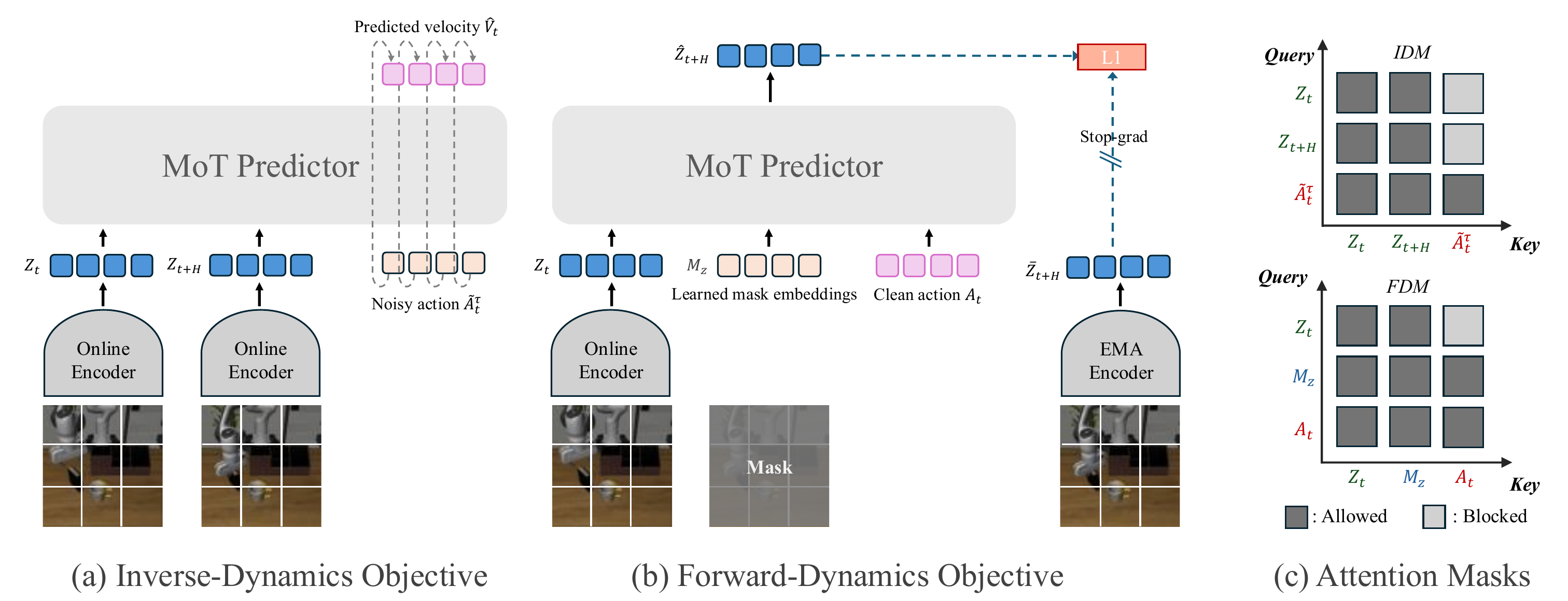}
    \caption{
        Action-grounded masked trajectory prediction.
        (a) Inverse dynamics predicts the flow velocity of noised actions conditioned on current and future observations.
        (b) Forward dynamics predicts masked future observation latents from the current observation and clean actions, supervised by a stop-gradient EMA target.
        (c) Objective-specific attention masks control information flow. Auxiliary tokens are omitted for clarity.
    }
    \label{fig:idm-fdm}
\end{figure}

\section{Method}
\label{sec:method}

SLIM learns language-conditioned manipulation as action-grounded interaction in observation-latent space.
The method separates three roles that are often entangled in large VLA or video-generative policies: language specifies the task, visual observations define the control state, and the action chunk specifies the continuous control trajectory to execute.
Rather than using a large VLM as the main action backbone or reconstructing future pixels, SLIM first learns how actions and observation latents explain one another, then trains a compact flow-matching policy for action generation.

\subsection{Problem Formulation}

Let $y$ denote the language instruction and let $\ell=g_{\mathrm{lang}}(y)$ be its encoded language condition, instantiated with T5-small \citep{raffel2020t5}.
At time $t$, the robot receives a multi-view visual observation $\mathbf{o}_t$ and proprioceptive state $\mathbf{q}_t$.
The visual encoder maps the observation to a sequence of $N_z$ observation-latent tokens
\begin{equation}
    \mathbf{Z}_t = g_{\psi}(\mathbf{o}_t)
    \in \mathbb{R}^{N_z \times d_z}.
    \label{eq:observation-latent}
\end{equation}
In our implementation, $g_{\psi}$ is initialized from DINOv2 \citep{oquab2023dinov2}.
The policy predicts an action chunk
\begin{equation}
    \mathbf{A}_t = (\mathbf{a}_t,\ldots,\mathbf{a}_{t+H-1}),
    \label{eq:action-chunk}
\end{equation}
where $H$ is the action horizon.
We use \emph{action tokens} to denote the continuous embeddings of the action chunk; no action discretization is applied.
The corresponding online future latent and its stable EMA target are
\begin{equation}
    \mathbf{Z}_{t+H}=g_{\psi}(\mathbf{o}_{t+H}),
    \qquad
    \bar{\mathbf{Z}}_{t+H}=g_{\bar{\psi}}(\mathbf{o}_{t+H}),
    \label{eq:future-latents}
\end{equation}
where $g_{\bar{\psi}}$ is an exponential moving average (EMA) copy of $g_{\psi}$.
The current and future online latents $\mathbf{Z}_t$ and $\mathbf{Z}_{t+H}$ therefore live in the same observation-latent space and differ only in temporal role.

\subsection{Latent Interaction Policy}

\begin{wrapfigure}{r}{0.40\textwidth}
    \vspace{-1.2em}
    \centering
    \includegraphics[width=0.38\textwidth]{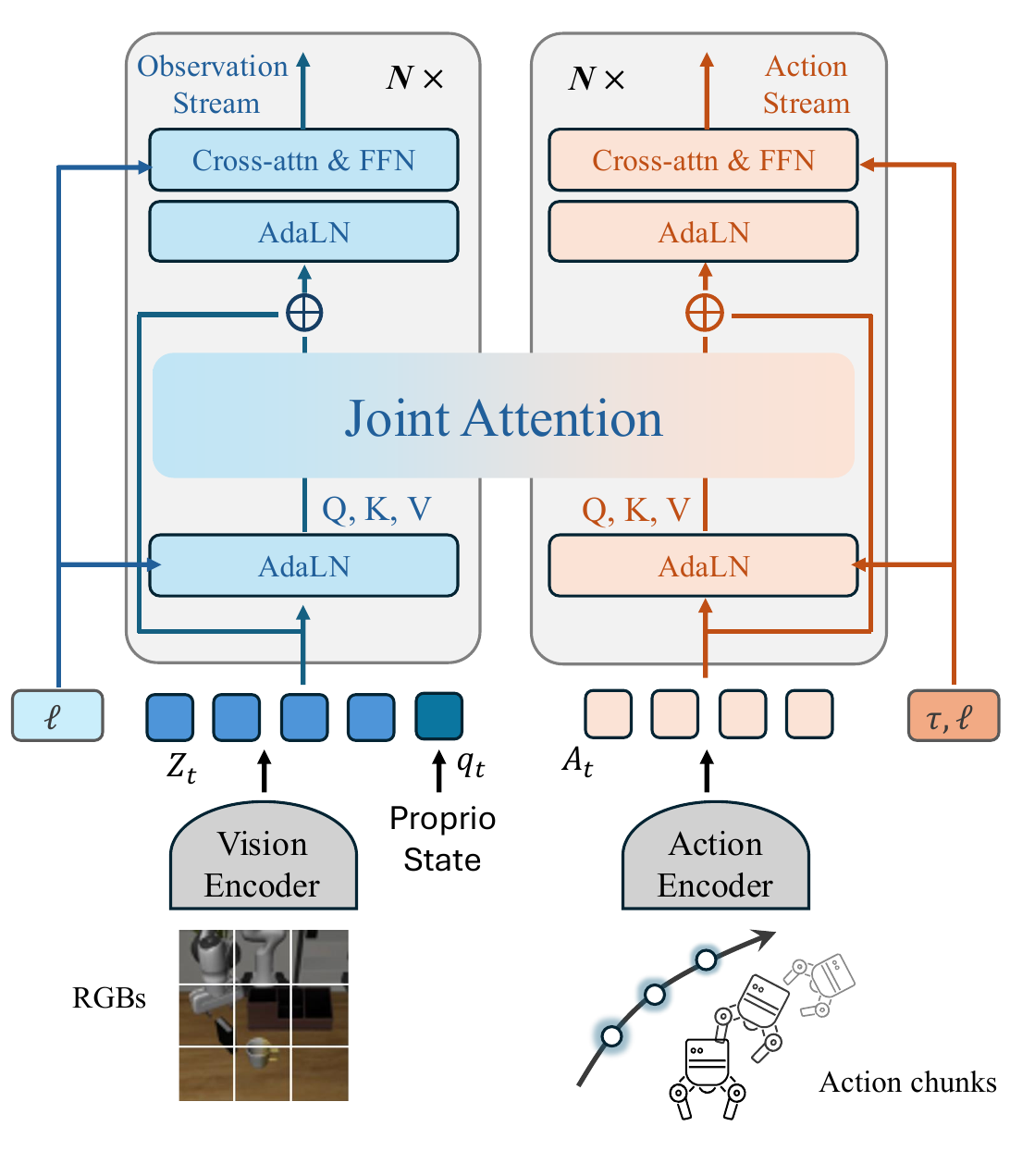}
    \caption{MoT backbone with observation-action joint attention and per-stream language cross-attention.}
    \label{fig:mot-architecture}
    \vspace{-1.0em}
\end{wrapfigure}

We instantiate the policy backbone as a compact Mixture-of-Transformers (MoT) \citep{liang2025mot}.
As illustrated in Figure~\ref{fig:mot-architecture}, MoT maintains two interacting streams: an observation stream and an action stream.
Before entering the MoT trunk, the inputs are mapped to its shared hidden dimension $d$ ($d=768$ in our implementation).
Each observation-latent token is linearly projected to $d$, and the proprioceptive state $\mathbf{q}_t$ is mapped by a separate linear projection to a single state token that is prepended to the observation stream.
For the action stream, a three-layer MLP encoder first projects each continuous action vector to $d$, concatenates it with a sinusoidal embedding of the flow timestep, and then maps the result to an action token in $\mathbb{R}^{d}$.
The observation stream is therefore grounded in both the current observation latent $\mathbf{Z}_t$ and proprioceptive state $\mathbf{q}_t$.
The observation stream also maintains $N_z$ learned future-slot embeddings, collected as
$\mathbf{M}_z=[\mathbf{m}_z^{(1)},\ldots,\mathbf{m}_z^{(N_z)}]^\top
\in\mathbb{R}^{N_z\times d}$.
They are shared across examples but distinct across future-token positions.
The future positions are filled with the projected clean future latent $\mathbf{Z}_{t+H}$ in the action-masked branch and with $\mathbf{M}_z$ in the future-latent-masked and policy branches.
The action stream carries the action variable used by the objective: the noised action chunk $\tilde{\mathbf{A}}_t^{\tau}$ for flow-matching action prediction, or the clean action chunk $\mathbf{A}_t$ for future-observation-latent prediction.
Language $\ell$ is injected through per-stream cross-attention rather than being treated as part of the joint observation-action stream.
We write the resulting action-velocity predictor as $v_{\theta}$ and the future-observation-latent predictor as $p_{\theta}$ in the objectives below.
This avoids assigning a single fixed action input to MoT: the concrete token contents are objective-specific, while the same two-stream interaction backbone is reused.

Within each MoT block, the two streams produce stream-specific queries, keys, and values, and then interact through a shared joint-attention operation.
The attention output is split back into observation-side and action-side tokens, followed by per-stream language cross-attention and feed-forward layers.
This design makes language a task condition while reserving the main computation for observation-action interaction.
During flow sampling, the observation latent and language context remain fixed, and the policy repeatedly updates the action stream to estimate the action velocity field.
Thus, SLIM behaves less like a large language-centric decoder and more like a compact interaction model over control-relevant latents and action tokens.

\subsection{Action-Grounded Training}

SLIM uses two training phases that share the same observation-action interaction backbone.
The first phase learns action-grounded predictive observation latents by reconstructing masked trajectory variables.
The second phase trains the final deployable policy with flow matching, using the current observation, proprioception, and language conditions, without requiring future observations at inference time.

\textbf{Stage-1 masked trajectory prediction.}
Stage 1 learns predictive observation latents by masking trajectory variables and reconstructing them from the remaining context, following the representation-space prediction principle of JEPA-style objectives \citep{assran2023ijepa,bardes2024vjepa}.
The supervision comes from robot trajectories themselves, without additional annotations or pixel reconstruction.
We use two complementary masked-prediction branches.

When the action chunk is masked, SLIM conditions on the current observation latent $\mathbf{Z}_t$, the clean future observation latent $\mathbf{Z}_{t+H}$, proprioception $\mathbf{q}_t$, and language $\ell$.
The masked action chunk is modeled with conditional flow matching \citep{lipman2023flowmatching}:
\begin{equation}
    \epsilon \sim \mathcal{N}(0,I),
    \qquad
    \tau \sim \mathcal{U}(0,1),
    \qquad
    \tilde{\mathbf{A}}_t^{\tau}
    =
    (1-\tau)\epsilon+\tau\mathbf{A}_t,
    \qquad
    \mathbf{V}_t^{\star}
    =
    \mathbf{A}_t-\epsilon.
    \label{eq:stage1-flow}
\end{equation}
The inverse-dynamics-style objective is
\begin{equation}
    \mathcal{L}_{\mathrm{IDM}}
    =
    \mathbb{E}
    \left[
    \left\|
    v_{\theta}(\tilde{\mathbf{A}}_t^{\tau},\tau
    \mid
    \mathbf{Z}_t,\mathbf{Z}_{t+H},\mathbf{q}_t,\ell)
    -
    \mathbf{V}_t^{\star}
    \right\|_2^2
    \right].
    \label{eq:idm-loss}
\end{equation}
This branch asks which action chunk explains the transition from $\mathbf{Z}_t$ to $\mathbf{Z}_{t+H}$.

When the future observation latent is masked, SLIM replaces its token positions with $\mathbf{M}_z$ and conditions on the current observation latent and clean action chunk.
The model predicts
\begin{equation}
    \hat{\mathbf{Z}}_{t+H}
    =
    p_{\theta}(\mathbf{M}_z
    \mid \mathbf{Z}_t,\mathbf{A}_t,\mathbf{q}_t,\ell),
    \label{eq:future-pred}
\end{equation}
and aligns it to the detached EMA target:
\begin{equation}
    \mathcal{L}_{\mathrm{FDM}}
    =
    \left\|
    \mathrm{LN}(\hat{\mathbf{Z}}_{t+H})
    -
    \mathrm{LN}(\mathrm{sg}(\bar{\mathbf{Z}}_{t+H}))
    \right\|_1 .
    \label{eq:fdm-loss}
\end{equation}
This forward-dynamics-style objective asks which future observation-latent factors are predictable from the current observation and action chunk.
The full Stage-1 loss is
\begin{equation}
    \mathcal{L}_{\mathrm{stage1}}
    =
    \lambda_{\mathrm{IDM}}\mathcal{L}_{\mathrm{IDM}}
    +
    \lambda_{\mathrm{FDM}}\mathcal{L}_{\mathrm{FDM}} .
    \label{eq:stage1-loss}
\end{equation}
Together, the two branches make the learned latent space action-grounded: actions must explain future observation latents, and future observation latents must constrain action reconstruction.
These inverse- and forward-dynamics-style signals are related to dynamics pretraining objectives for robot learning \citep{disentangled_dynamics2026}.

\textbf{Stage-2 flow-matching policy training.}
After masked trajectory prediction, SLIM trains the final policy for action generation.
This training setup matches inference: the policy conditions on the current observation latent $\mathbf{Z}_t$, learned future-slot embeddings $\mathbf{M}_z$, proprioception $\mathbf{q}_t$, language $\ell$, and the noised action chunk used by flow sampling.
No future observation latent is provided to the policy at inference time.
Using the same flow interpolation as in Eq.~\ref{eq:stage1-flow}, the Stage-2 objective is
\begin{equation}
    \mathcal{L}_{\mathrm{FM}}
    =
    \mathbb{E}
    \left[
    \left\|
    v_{\theta}(\tilde{\mathbf{A}}_t^{\tau},\tau
    \mid
    \mathbf{Z}_t,\mathbf{M}_z,\mathbf{q}_t,\ell)
    -
    \mathbf{V}_t^{\star}
    \right\|_2^2
    \right].
    \label{eq:stage2-fm}
\end{equation}
The training objective in Stage 2 is simply
\begin{equation}
    \mathcal{L}_{\mathrm{stage2}}
    =
    \mathcal{L}_{\mathrm{FM}} .
    \label{eq:stage2-loss}
\end{equation}
Thus, Stage 1 uses future observation latents to learn action-grounded transition structure.
In Stage 2, future observation latents are no longer given as inputs or supervised as explicit targets.
However, the observation-side hidden states still contain the predictive slots shaped by Stage-1 masked trajectory prediction, and the action stream attends to these hidden states through MoT joint attention during flow sampling.
Action generation during inference is therefore guided by an implicit future-latent prediction structure, rather than by an explicit future-observation-latent prediction. This makes SLIM computationally efficient at inference time.

\section{Experiments}
\label{sec:experiments}

\begin{figure*}[t]
\centering
\includegraphics[width=\textwidth]{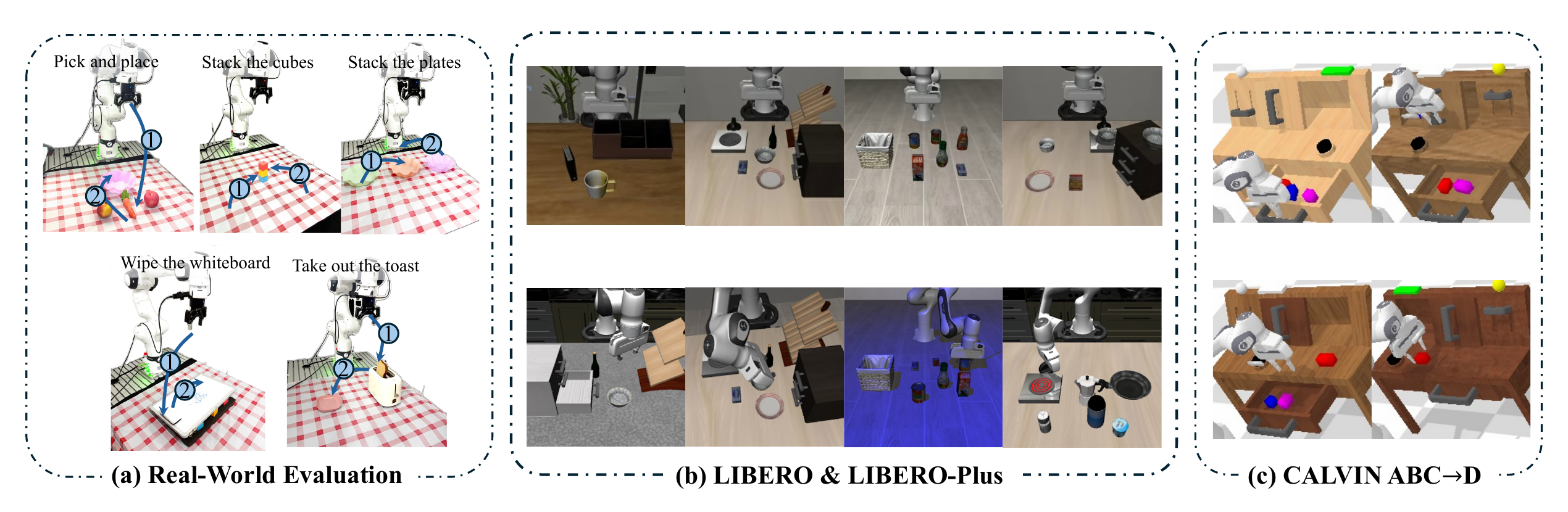}
\caption{
Overview of the evaluation settings.
(a) Representative physical manipulation tasks used in the real-world evaluation.
(b) LIBERO and LIBERO-Plus evaluate language-conditioned manipulation and robustness under controlled perturbations.
(c) CALVIN ABC$\rightarrow$D evaluates long-horizon task composition in an unseen environment.
}
\label{fig:experiment-settings}
\end{figure*}

We evaluate SLIM in simulation and on real-world tasks.
The simulation study tests nominal language-conditioned manipulation, robustness under controlled distribution shifts, and long-horizon composition in an unseen environment.
The real-world study is designed to test whether the same compact latent interaction policy can be trained and executed on physical manipulation tasks.
Figure~\ref{fig:experiment-settings} summarizes the evaluation settings.

\subsection{Simulation Evaluation}

\paragraph{Benchmarks.}
We use three complementary simulation benchmarks.
LIBERO evaluates language-conditioned manipulation on four standard task suites: LIBERO-10, LIBERO-Object, LIBERO-Spatial, and LIBERO-Goal \citep{liu2023libero}.
LIBERO-Plus builds controlled perturbations on top of LIBERO, including camera, robot initial state, language, lighting, background, sensor noise, and object-layout shifts \citep{fei2025liberoplus}.
CALVIN ABC$\rightarrow$D evaluates long-horizon language-conditioned task composition: the policy is trained on environments A/B/C and evaluated in the unseen environment D, where each evaluation chain contains up to five sequential language instructions \citep{mees2021calvin}.
We report success rate for LIBERO and LIBERO-Plus, and average sequence length for CALVIN.

\paragraph{Baselines.}
We compare against representative vision-language-action (VLA) policies and WAM-style methods.
The combined LIBERO and LIBERO-Plus table and the CALVIN table include model size and a P.T. indicator, where P.T. denotes additional embodied policy/world pretraining before target-suite training; generic visual-language or video backbone initialization alone is not counted.
The LIBERO-family comparisons include OpenVLA-OFT~\citep{kim2025fine}, NORA~\citep{hung2025nora}, WorldVLA~\citep{cen2025worldvla}, UnifiedVLA~\citep{wang2026unified}, and RIPT-VLA~\citep{tan2025interactive}.
The CALVIN comparisons further include RoboFlamingo~\citep{li2024vision}, RoboDual~\citep{bu2024towards}, FLOWER~\citep{reuss2025flower}, Seer-Large~\citep{tian2025predictive}, GR-MG~\citep{li2025gr}, UniPi~\citep{du2023learning}, and VPP~\citep{hu2024video}.
For readability, the main table reports a compact subset of baselines while preserving representative VLA and WAM comparisons; detailed original-LIBERO suite results are provided in Appendix Table~\ref{tab:libero-original-baselines}.

\paragraph{Training and testing setting.}
For LIBERO and LIBERO-Plus, SLIM is trained only on the original LIBERO data.
We first run Stage-1 masked trajectory learning to make the observation latents action-grounded, and then train the final flow-matching policy.
LIBERO-Plus is evaluated in a zero-shot setting: the same checkpoint trained on original LIBERO is directly tested on LIBERO-Plus without using LIBERO-Plus training data.
For CALVIN, SLIM is trained on the ABC training environments and evaluated on the held-out D environment.

\paragraph{Results.}
Tables~\ref{tab:libero-plus-baselines} and~\ref{tab:calvin-baselines} summarize the simulation results, with detailed original-LIBERO results deferred to Appendix Table~\ref{tab:libero-original-baselines}.
SLIM reaches \textbf{97.5\%} overall success on original LIBERO.
On LIBERO-Plus, the same LIBERO-trained checkpoint obtains \textbf{77.45\%} overall success under zero-shot testing, with strong performance on camera, language, lighting, noise, and layout perturbations.
On CALVIN ABC$\rightarrow$D, SLIM obtains \textbf{4.556} average sequence length, indicating strong long-horizon task composition in an unseen environment.

\begin{table*}[!htbp]
\caption{Results on LIBERO and LIBERO-Plus (\%). LIBERO reports the original-benchmark overall score, while LIBERO-Plus evaluates zero-shot robustness under seven perturbations. P.T. denotes additional embodied policy/world pretraining, excluding generic vision-language or video initialization.}
\label{tab:libero-plus-baselines}
\centering
\small
\setlength{\tabcolsep}{1.7pt}
\renewcommand{\arraystretch}{1.08}
\begin{tabular}{lcc@{\hspace{2pt}}!{\vrule width 0.4pt}@{\hspace{2pt}}c@{\hspace{2pt}}!{\vrule width 0.4pt}@{\hspace{2pt}}ccccccc@{\hspace{2pt}}!{\vrule width 0.4pt}@{\hspace{2pt}}c}
\gapthickhline
Method & P.T. & Size & LIBERO & Camera & Robot & Language & Light & Background & Noise & Layout & Overall \\
\gapthickhline
\multicolumn{12}{l}{\textit{VLA baselines}} \\
\gaphline
OpenVLA & \cmark & 7B & 76.5 & 0.8 & 3.5 & 23.0 & 8.1 & 34.8 & 15.2 & 28.5 & 15.6 \\
OpenVLA-OFT & \cmark & 7B & 97.1 & 56.4 & 31.9 & \third{79.5} & 88.7 & \second{93.3} & \third{75.8} & 74.3 & \third{69.6} \\
$\pi_0$ & \cmark & 3.3B & 94.1 & \third{61.0} & 40.8 & 63.7 & \third{89.3} & 84.1 & \second{80.1} & \third{75.9} & 69.3 \\
NORA & \cmark & 3B & 87.9 & 2.2 & 37.0 & 65.1 & 45.7 & 58.6 & 12.8 & 62.1 & 39.0 \\
WorldVLA & \xmark & 7B & 79.1 & 0.1 & 27.9 & 41.6 & 43.7 & 17.1 & 11.0 & 38.0 & 25.0 \\
UnifiedVLA & \cmark & 8.5B & 95.5 & 1.8 & \second{46.2} & 69.5 & 69.0 & 81.0 & 21.2 & 31.9 & 42.9 \\
RIPT-VLA & \cmark & 7B & \second{97.5} & 55.2 & 31.2 & 77.6 & 88.4 & 91.6 & 73.5 & 74.2 & 68.4 \\
\gapthickhline
\multicolumn{12}{l}{\textit{WAM baselines}} \\
\gaphline
Fast-WAM & \xmark & 6B & \best{97.6} & 16.4 & \third{44.5} & 68.9 & 78.2 & 53.7 & 37.7 & 60.7 & 50.0 \\
VLA-JEPA & \cmark & 3B & \third{97.2} & \second{64.2} & \best{67.7} & \best{88.1} & \second{91.8} & \best{93.4} & 65.8 & \best{83.9} & \best{79.5} \\
\gapthickhline
\multicolumn{12}{l}{\textit{Ours}} \\
\gaphline
\textbf{SLIM} & \xmark & \textbf{0.47B} & \second{97.5} & \best{70.73} & 36.90 & \second{87.57} & \best{94.75} & \third{92.01} & \best{86.07} & \second{83.21} & \second{77.45} \\
\gapthickhline
\end{tabular}%
\end{table*}

Table~\ref{tab:calvin-baselines} reports CALVIN ABC$\rightarrow$D.
Unlike LIBERO, this benchmark uses average sequence length rather than success rate.
SLIM achieves 4.556 out of 5 on the unseen-environment split, suggesting that the learned latent interaction policy supports long-horizon language-conditioned task composition beyond the LIBERO family.

\begin{table*}[t]
\caption{Results on CALVIN ABC$\rightarrow$D. Task 1--5 report sequential success rates (\%), and Avg. Length is the primary metric. P.T. denotes additional embodied policy/world pretraining. Model sizes are reported in billions of parameters (B); ``--'' indicates unavailable data.}
\label{tab:calvin-baselines}
\centering
\small
\setlength{\tabcolsep}{2.0pt}
\renewcommand{\arraystretch}{1.08}
\begin{tabular}{lccccccc@{\hspace{3pt}}!{\vrule width 0.4pt}@{\hspace{3pt}}c}
\gapthickhline
Method & P.T. & Size & Task 1 & Task 2 & Task 3 & Task 4 & Task 5 & Avg. Length \\
\gapthickhline
\multicolumn{9}{l}{\textit{VLA baselines}} \\
\gaphline
RoboFlamingo & \xmark & 3B & 82.4 & 61.9 & 46.6 & 33.1 & 23.5 & 2.47 \\
OpenVLA & \cmark & 7B & 91.3 & 77.8 & 62.0 & 52.1 & 43.5 & 3.27 \\
RoboDual & \cmark & 7.02B & 94.4 & 82.7 & 72.1 & 62.4 & 54.4 & 3.66 \\
UnifiedVLA & \cmark & 8.5B & \third{98.9} & \third{94.8} & 89.0 & 82.8 & 75.1 & 4.41 \\
FLOWER & \cmark & 0.95B & \best{99.4} & \second{95.8} & \second{90.7} & \second{84.9} & \third{77.8} & \second{4.53} \\
Seer-Large & \cmark & 0.57B & 96.3 & 91.6 & 86.1 & 80.3 & 74.0 & 4.28 \\
GR-MG & \xmark & -- & 96.8 & 89.3 & 81.5 & 72.7 & 64.4 & 4.04 \\
\gapthickhline
\multicolumn{9}{l}{\textit{WAM baselines}} \\
\gaphline
GR-1 & \xmark & 0.20B & 85.4 & 71.2 & 59.6 & 49.7 & 40.1 & 3.06 \\
UniPi & \xmark & -- & 56.0 & 16.0 & 8.0 & 8.0 & 4.0 & 0.92 \\
VPP & \cmark & 1.5B & 96.5 & 90.9 & 86.6 & 82.0 & 76.9 & 4.33 \\
DreamVLA & \cmark & -- & 98.2 & 94.6 & \third{89.5} & \third{83.4} & \second{78.1} & \third{4.44} \\
\gapthickhline
\multicolumn{9}{l}{\textit{Ours}} \\
\gaphline
\textbf{SLIM} & \xmark & \textbf{0.47B} & \second{99.3} & \best{96.7} & \best{92.3} & \best{87.1} & \best{80.2} & \best{4.556} \\
\gapthickhline
\end{tabular}
\end{table*}

\subsection{Real-World Evaluation}

We further evaluate SLIM on a real-world manipulation suite.
The goal of this study is to test whether the compact latent interaction policy remains effective when trained from physical demonstrations and evaluated under visual and scene-level disturbances.
We compare SLIM with two representative baselines, $\pi_{0.5}$ and Fast-WAM, which respectively represent a strong VLA policy and a WAM-style policy.

\paragraph{Data and training.}
We design five real-world manipulation tasks: placing a carrot into a bowl, stacking three plates, taking a piece of toast out of a toaster and placing it onto a plate, stacking blocks, and wiping a whiteboard.
For each task, we collect 150 demonstrations, resulting in 750 demonstrations in total.
All tasks are mixed during training, so each policy is trained as a multi-task real-world policy rather than as five separately trained task-specific policies.
SLIM follows the same two-stage recipe as in simulation: Stage-1 masked trajectory learning is applied to the collected robot trajectories, followed by flow-matching policy training for action generation.
The baselines are trained on the same demonstration mixture for a controlled comparison.

\begin{wrapfigure}{r}{0.50\textwidth}
    \vspace{-1.2em}
    \centering
    \includegraphics[width=0.48\textwidth]{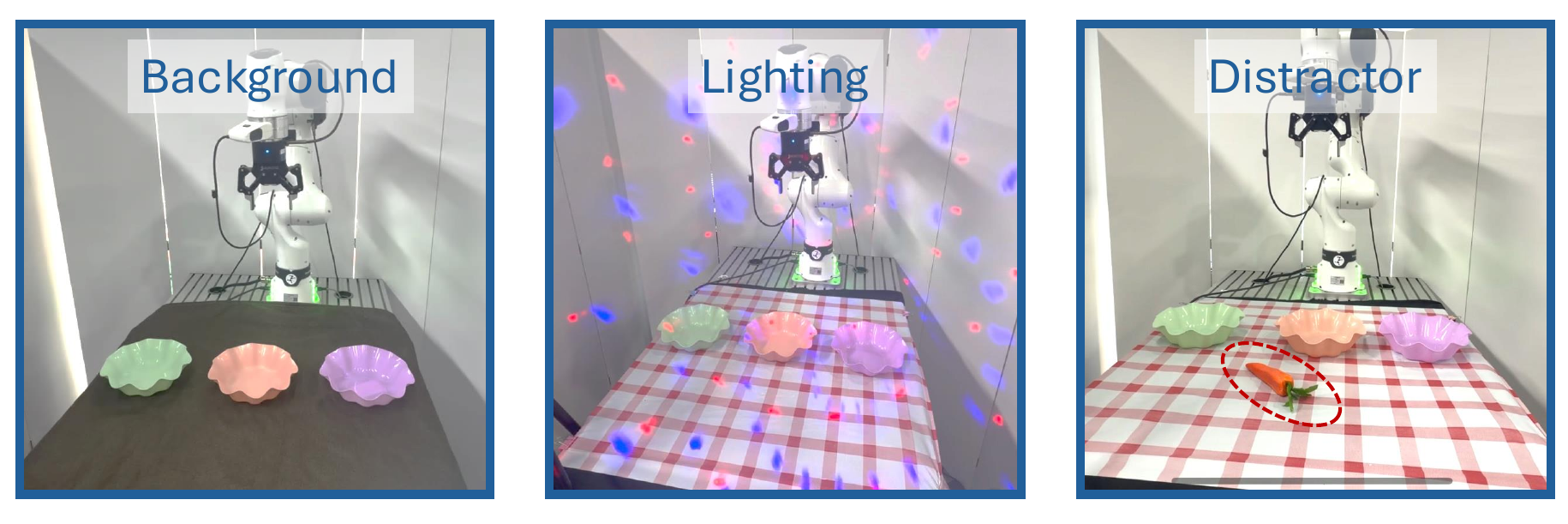}
    \caption{Representative real-world OOD settings: distractor objects, background texture, and lighting.}
    \label{fig:real-world-ood-settings}
    \vspace{-1.0em}
\end{wrapfigure}

\paragraph{Evaluation protocol.}
We evaluate each policy on the five real-world tasks under the nominal setting and three perturbation dimensions: background texture changes, lighting changes, and added distractor objects.
Each task-condition pair is evaluated with 10 trials.
This protocol measures both task completion under the training-like setup and robustness to common real-world visual shifts.
Representative perturbations are shown in Figure~\ref{fig:real-world-ood-settings}, while the complete task-specific OOD configurations are provided in Appendix Figure~\ref{fig:real-world-ood-details}.

\begin{figure*}[t]
\centering
\includegraphics[width=0.92\textwidth]{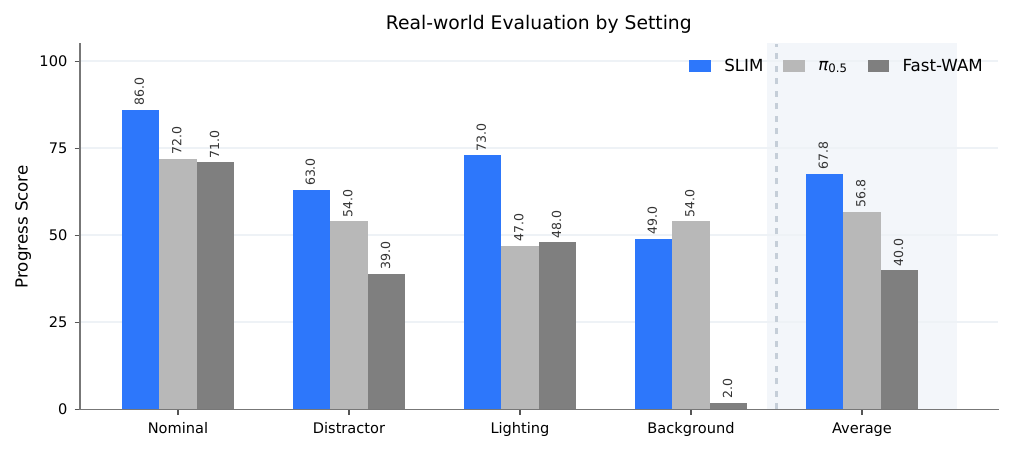}
\caption{Average real-world progress by evaluation setting. Each setting averages the five tasks, and the final group averages all four settings. Task-wise results and scoring milestones are provided in Appendix~\ref{app:real-world-details}.}
\label{fig:real-world-progress}
\end{figure*}

\paragraph{Results.}
We report progress scores across tasks and perturbation dimensions, comparing SLIM against $\pi_{0.5}$ and Fast-WAM.
Each trial receives task-dependent partial credit, and the reported value is 100 times the mean progress over 10 trials; the 0.5 and 1.0 milestones for every task are defined in Appendix~\ref{app:real-world-details}.
As shown in Figure~\ref{fig:real-world-progress}, SLIM obtains the highest average progress under the nominal, distractor, and lighting settings.
Under the more challenging background shift, SLIM achieves an average progress of 49, close to $\pi_{0.5}$ at 54, while Fast-WAM drops to 2.
The complete task-wise breakdown for every setting is provided in Appendix Figure~\ref{fig:real-world-progress-detailed}.

\subsection{Ablation Study}

\begin{figure*}[t]
\centering
\includegraphics[width=\textwidth]{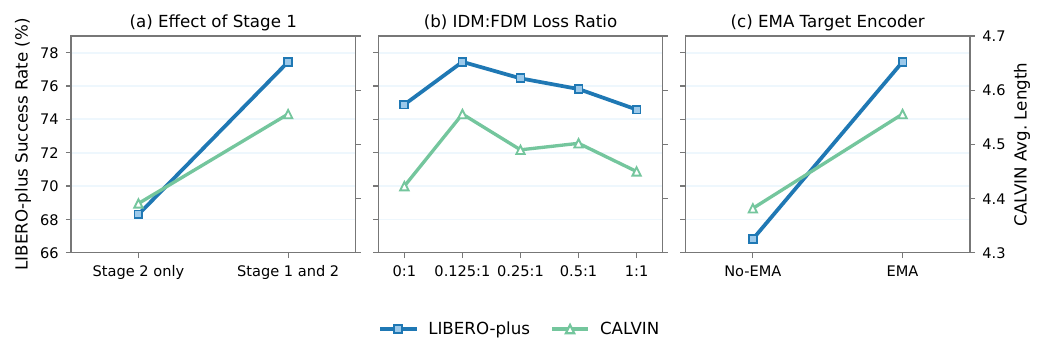}
\caption{Ablations of the action-grounded masked-prediction pipeline. We compare Stage-2-only training against the full two-stage recipe, vary the IDM:FDM loss ratio, and evaluate the EMA target encoder on LIBERO-Plus and CALVIN.}
\label{fig:ablation-lines}
\end{figure*}

We organize the ablations around three questions: whether Stage-1 masked trajectory learning improves the final policy, how sensitive performance is to the IDM:FDM loss ratio, and how the EMA target encoder affects downstream performance.
Figure~\ref{fig:ablation-lines} summarizes these comparisons on LIBERO-Plus and CALVIN.
Adding Stage 1 consistently improves both benchmarks over Stage-2-only training.
Across loss ratios, an IDM:FDM weight of $0.125{:}1$ provides the strongest joint result; performance remains relatively stable over moderate ratios and degrades toward the extremes.
Finally, the EMA target improves LIBERO-Plus success from 66.82\% to 77.45\% and CALVIN average sequence length from 4.382 to 4.556, showing that a slowly moving target is important for stable predictive representation learning.

Table~\ref{tab:ema-collapse} further examines the EMA target through representation diagnostics.
Removing the EMA target causes a clear loss of representation diversity: over the final 20 evaluation probes, the future-latent effective rank decreases from 61.28 to 13.95, while token cosine similarity increases from 0.071 to 0.352.
Although future-latent MSE decreases from 0.245 to 0.166, the lower error reflects a degenerate low-rank target rather than improved prediction: top-1 energy concentration rises from 0.097 to 0.395 and top-5 concentration from 0.362 to 0.674.
This degradation is accompanied by a drop in LIBERO-Plus success from 77.45\% to 66.82\%, supporting the view that EMA preserves latent diversity and closed-loop robustness.

\begin{table}[t]
\caption{EMA target ablation and collapse diagnostics averaged over the final 20 Stage-1 evaluation probes. Higher effective rank and lower energy concentration/cosine indicate more diverse latent usage.}
\label{tab:ema-collapse}
\centering
\small
\setlength{\tabcolsep}{2.5pt}
\renewcommand{\arraystretch}{1.08}
\begin{tabular}{lccccc}
\gapthickhline
Variant & Eff. rank $\uparrow$ & Top-1 $\downarrow$ & Top-5 $\downarrow$ & Cosine $\downarrow$ & Future-latent MSE $\downarrow$ \\
\gapthickhline
EMA & 61.28 & 0.097 & 0.362 & 0.071 & 0.245 \\
No EMA & 13.95 & 0.395 & 0.674 & 0.352 & 0.166 \\
\gapthickhline
\end{tabular}
\end{table}

\subsection{Analysis}

\begin{figure*}[t]
\centering
\includegraphics[width=\textwidth]{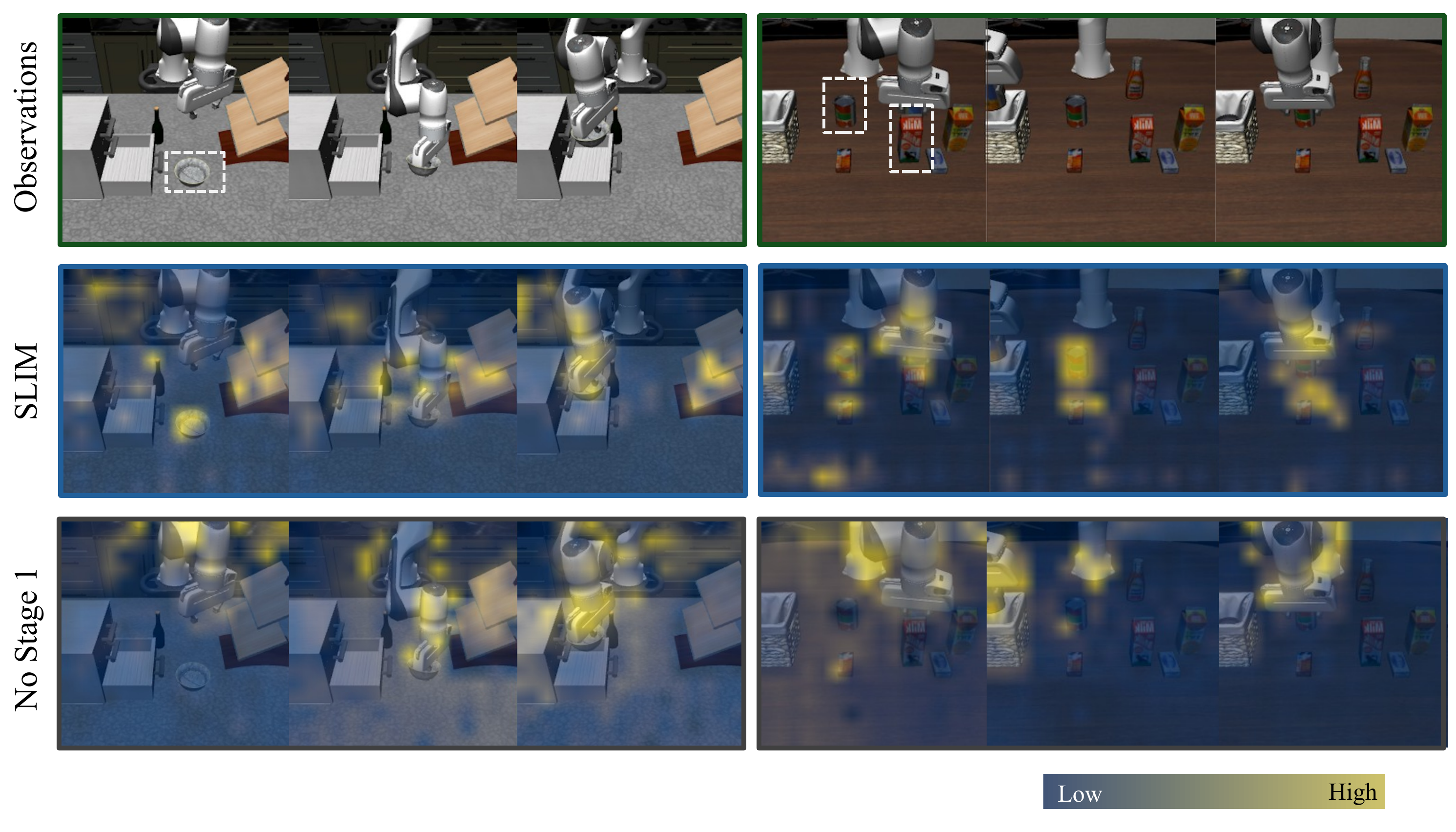}
\caption{Action-grounding analysis across two representative manipulation sequences. The top row shows observations, with task-relevant objects indicated by dashed boxes; the middle and bottom rows show action-to-observation attention for SLIM and the step-matched policy without Stage 1. Warmer colors denote larger attention. SLIM more consistently follows the manipulated object, gripper, and their interaction region as the task progresses.}
\label{fig:action-grounding-analysis}
\end{figure*}

We first examine whether Stage-1 masked trajectory learning changes the visual evidence used by the action stream, and then quantify the efficiency of the resulting policy.
The attention probe compares the full SLIM model with the policy trained without Stage 1, as visualized in Figure~\ref{fig:action-grounding-analysis}.

\paragraph{Are the observation latents action-grounded?}
An action-grounded latent should expose visual factors that matter for the current manipulation rather than preserve only generic visual similarity.
We extract action-to-observation attention from the MoT backbone and project the patch-level scores back to the image plane.
Figure~\ref{fig:action-grounding-analysis} shows two representative sequences at three manipulation phases.
As the interaction progresses, SLIM more consistently follows the task-relevant object, the gripper, and their contact region.
Without Stage 1, attention is more diffuse and more often dominated by the robot body or unrelated scene regions.
The maps are qualitative probes rather than causal attributions; together with the Stage-1 ablation in Figure~\ref{fig:ablation-lines}, they indicate that masked trajectory learning shifts the visual evidence used by the action stream toward manipulation-relevant structure.

\paragraph{Latency, memory, and computation.}
Table~\ref{tab:latency-flops} reports controlled model-inference latency, peak VRAM, and per-chunk FLOPs on a single NVIDIA H100 80GB.
The protocol, model-specific inputs, and FLOPs methodology are provided in Appendix~\ref{app:inference-benchmark}.

\begin{table}[t]
\caption{Controlled policy inference on one NVIDIA H100 80GB. Latency is mean model-inference time, peak VRAM is incremental device memory, and FLOPs are reported per action chunk. All models use PyTorch eager and BF16.}
\label{tab:latency-flops}
\centering
\small
\setlength{\tabcolsep}{3.5pt}
\renewcommand{\arraystretch}{1.08}
\begin{tabular}{lrrr}
\gapthickhline
Model & Latency (ms) $\downarrow$ & Peak VRAM (GiB) $\downarrow$ & GFLOPs/chunk $\downarrow$ \\
\gapthickhline
Fast-WAM & 360.6 & 13.63 & 2090.07 \\
$\pi_{0.5}$ & 193.1 & 7.94 & 4714.59 \\
\textbf{SLIM} & \textbf{60.6} & \textbf{4.26} & \textbf{490.73} \\
\gapthickhline
\end{tabular}
\end{table}

SLIM averages 60.6 ms per inference call, making it $3.19\times$ faster than $\pi_{0.5}$ and $5.95\times$ faster than Fast-WAM under the controlled PyTorch-eager protocol.
Its pooled p95 latency is 61.6 ms over 600 synchronized calls.
SLIM uses 4.26 GiB of peak incremental GPU memory, $1.86\times$ less than $\pi_{0.5}$ and $3.20\times$ less than Fast-WAM.
The PyTorch profiler reports $9.61\times$ fewer FLOPs per action chunk than $\pi_{0.5}$ and $4.26\times$ fewer than Fast-WAM under their native action horizons and sampling schedules.

\section{Conclusion}
\label{sec:conclusion}

We introduced SLIM, a compact latent interaction policy that learns control-relevant structure directly from robot trajectories.
Its action-grounded masked trajectory objective couples inverse dynamics, which reconstructs actions from current and future observation latents, with forward dynamics, which predicts future observation latents from current observations and actions.
This bidirectional training signal shapes a compact MoT backbone without requiring a large VLM for action generation or pixel-level future prediction at inference time.
Across LIBERO, zero-shot LIBERO-Plus, CALVIN ABC$\rightarrow$D, and real-world manipulation, SLIM is competitive with or stronger than representative VLA and world-action-model baselines while using fewer parameters and achieving lower inference latency and GPU memory usage.
The ablations attribute the gains to masked trajectory learning, while the qualitative attention maps suggest a stronger focus on manipulation-relevant objects and interaction regions.

Our experiments focus on a single compact model scale and therefore do not establish how this paradigm scales with model capacity, pretraining data, or embodiment diversity.
Systematically characterizing these scaling axes is an important direction for future work.
In particular, scaling action-grounded latent interaction with broader cross-embodiment trajectories may combine the efficiency demonstrated here with the additional capacity needed for more diverse tasks and environments.

\bibliography{references}

@article{brohan2022rt1,
  title={{RT-1}: Robotics Transformer for Real-World Control at Scale},
  author={Brohan, Anthony and Brown, Noah and Carbajal, Justice and Chebotar, Yevgen and Dabis, Joseph and Finn, Chelsea and Gopalakrishnan, Keerthana and Hausman, Karol and Herzog, Alex and Hsu, Jasmine and others},
  journal={arXiv preprint arXiv:2212.06817},
  year={2022}
}

@inproceedings{brohan2023rt2,
  title={{RT-2}: Vision-Language-Action Models Transfer Web Knowledge to Robotic Control},
  author={Zitkovich, Brianna and Yu, Tianhe and Xu, Sichun and Xu, Peng and Xiao, Ted and Xia, Fei and Wu, Jialin and Wohlhart, Paul and Welker, Stefan and Wahid, Ayzaan and others},
  booktitle={Conference on Robot Learning},
  pages={2165--2183},
  year={2023},
  organization={PMLR}
}

@article{driess2023palme,
  title={{PaLM-E}: An Embodied Multimodal Language Model},
  author={Driess, Danny and Xia, Fei and Sajjadi, Mehdi SM and Lynch, Corey and Chowdhery, Aakanksha and Ichter, Brian and Wahid, Ayzaan and Tompson, Jonathan and Vuong, Quan and Yu, Tianhe and others},
  journal={arXiv preprint arXiv:2303.03378},
  year={2023}
}

@article{openx2023rtx,
  title={Open X-Embodiment: Robotic Learning Datasets and {RT-X} Models},
  author={{Open X-Embodiment Collaboration} and O'Neill, Abby and Rehman, Abdul and Gupta, Abhinav and Maddukuri, Abhiram and Gupta, Abhishek and Padalkar, Abhishek and Lee, Abraham and Pooley, Acorn and Gupta, Agrim and others},
  journal={arXiv preprint arXiv:2310.08864},
  year={2023}
}

@article{octo2024,
  title={Octo: An open-source generalist robot policy},
  author={{Octo Model Team} and Ghosh, Dibya and Walke, Homer and Pertsch, Karl and Black, Kevin and Mees, Oier and Dasari, Sudeep and Hejna, Joey and Kreiman, Tobias and Xu, Charles and others},
  journal={arXiv preprint arXiv:2405.12213},
  year={2024}
}

@misc{kim2024openvla,
      title={OpenVLA: An Open-Source Vision-Language-Action Model}, 
      author={Moo Jin Kim and Karl Pertsch and Siddharth Karamcheti and Ted Xiao and Ashwin Balakrishna and Suraj Nair and Rafael Rafailov and Ethan Foster and Grace Lam and Pannag Sanketi and Quan Vuong and Thomas Kollar and Benjamin Burchfiel and Russ Tedrake and Dorsa Sadigh and Sergey Levine and Percy Liang and Chelsea Finn},
      year={2024},
      eprint={2406.09246},
      archivePrefix={arXiv},
      primaryClass={cs.RO},
      url={https://arxiv.org/abs/2406.09246}, 
}

@inproceedings{kim2025fine,
  title={Fine-Tuning Vision-Language-Action Models: Optimizing Speed and Success},
  author={Kim, Moo Jin and Finn, Chelsea and Liang, Percy},
  booktitle={Robotics: Science and Systems},
  year={2025},
  url={https://roboticsconference.org/2025/program/papers/17/}
}

@article{hung2025nora,
  title={{NORA}: A Small Open-Sourced Generalist Vision Language Action Model for Embodied Tasks},
  author={Hung, Chia-Yu and Sun, Qi and Hong, Pengfei and Zadeh, Amir and Li, Chuan and Tan, U-Xuan and Majumder, Navonil and Poria, Soujanya},
  journal={arXiv preprint arXiv:2504.19854},
  year={2025}
}

@article{cen2025worldvla,
  title={{WorldVLA}: Towards Autoregressive Action World Model},
  author={Cen, Jun and Yu, Chaohui and Yuan, Hangjie and Jiang, Yuming and Huang, Siteng and Guo, Jiayan and Li, Xin and Song, Yibing and Luo, Hao and Wang, Fan and Zhao, Deli and Chen, Hao},
  journal={arXiv preprint arXiv:2506.21539},
  year={2025}
}

@inproceedings{wang2026unified,
  title={Unified Vision-Language-Action Model},
  author={Wang, Yuqi and Li, Xinghang and Wang, Wenxuan and Zhang, Junbo and Li, Yingyan and Chen, Yuntao and Wang, Xinlong and Zhang, Zhaoxiang},
  booktitle={International Conference on Learning Representations},
  year={2026}
}

@inproceedings{tan2025interactive,
  title={Interactive Post-Training for Vision-Language-Action Models},
  author={Tan, Shuhan and Dou, Kairan and Zhao, Yue and Kr{\"a}henb{\"u}hl, Philipp},
  booktitle={International Conference on Learning Representations},
  year={2026}
}

@misc{black2024pi0,
      title={{$\pi_0$}: A Vision-Language-Action Flow Model for General Robot Control},
      author={Kevin Black and Noah Brown and Danny Driess and Adnan Esmail and Michael Equi and Chelsea Finn and Niccolo Fusai and Lachy Groom and Karol Hausman and Brian Ichter and Szymon Jakubczak and Tim Jones and Liyiming Ke and Sergey Levine and Adrian Li-Bell and Mohith Mothukuri and Suraj Nair and Karl Pertsch and Lucy Xiaoyang Shi and James Tanner and Quan Vuong and Anna Walling and Haohuan Wang and Ury Zhilinsky},
      year={2024},
      eprint={2410.24164},
      archivePrefix={arXiv},
      primaryClass={cs.LG},
      url={https://arxiv.org/abs/2410.24164}, 
}

@article{pi05_2025,
  title={{$\pi_{0.5}$}: A Vision-Language-Action Model with Open-World Generalization},
  author={{Physical Intelligence} and Black, Kevin and Brown, Noah and Darpinian, James and Dhabalia, Karan and Driess, Danny and Esmail, Adnan and Equi, Michael and Finn, Chelsea and Fusai, Niccolo and others},
  journal={arXiv preprint arXiv:2504.16054},
  year={2025}
}

@misc{grootn1_2025,
      title={GR00T N1: An Open Foundation Model for Generalist Humanoid Robots}, 
      author={{NVIDIA} and Johan Bjorck and Fernando Castañeda and Nikita Cherniadev and Xingye Da and Runyu Ding and Linxi {Jim} Fan and Yu Fang and Dieter Fox and Fengyuan Hu and Spencer Huang and Joel Jang and Zhenyu Jiang and Jan Kautz and Kaushil Kundalia and Lawrence Lao and Zhiqi Li and Zongyu Lin and Kevin Lin and Guilin Liu and Edith Llontop and Loic Magne and Ajay Mandlekar and Avnish Narayan and Soroush Nasiriany and Scott Reed and You Liang Tan and Guanzhi Wang and Zu Wang and Jing Wang and Qi Wang and Jiannan Xiang and Yuqi Xie and Yinzhen Xu and Zhenjia Xu and Seonghyeon Ye and Zhiding Yu and Ao Zhang and Hao Zhang and Yizhou Zhao and Ruijie Zheng and Yuke Zhu},
      year={2025},
      eprint={2503.14734},
      archivePrefix={arXiv},
      primaryClass={cs.RO},
      url={https://arxiv.org/abs/2503.14734}, 
}

@article{liang2025mot,
  title={Mixture-of-Transformers: A Sparse and Scalable Architecture for Multi-Modal Foundation Models},
  author={Liang, Weixin and Yu, Lili and Luo, Liang and Iyer, Srini and Dong, Ning and Zhou, Chunting and Ghosh, Gargi and Lewis, Mike and Yih, Wen-tau and Zettlemoyer, Luke and others},
  journal={Transactions on Machine Learning Research},
  year={2024}
}

@article{oquab2023dinov2,
  title={{DINOv2}: Learning Robust Visual Features without Supervision},
  author={Oquab, Maxime and Darcet, Timoth{\'e}e and Moutakanni, Th{\'e}o and Vo, Huy and Szafraniec, Marc and Khalidov, Vasil and Fernandez, Pierre and Haziza, Daniel and Massa, Francisco and El-Nouby, Alaaeldin and others},
  journal={Transactions on Machine Learning Research},
  year={2024}
}

@article{raffel2020t5,
  title={Exploring the Limits of Transfer Learning with a Unified Text-to-Text Transformer},
  author={Raffel, Colin and Shazeer, Noam and Roberts, Adam and Lee, Katherine and Narang, Sharan and Matena, Michael and Zhou, Yanqi and Li, Wei and Liu, Peter J.},
  journal={Journal of Machine Learning Research},
  volume={21},
  number={140},
  pages={1--67},
  year={2020}
}

@article{chi2023diffusionpolicy,
  title={Diffusion policy: Visuomotor policy learning via action diffusion},
  author={Chi, Cheng and Xu, Zhenjia and Feng, Siyuan and Cousineau, Eric and Du, Yilun and Burchfiel, Benjamin and Tedrake, Russ and Song, Shuran},
  journal={The International Journal of Robotics Research},
  volume={44},
  number={10-11},
  pages={1684--1704},
  year={2025},
  publisher={Sage Publications Sage UK: London, England}
}

@inproceedings{liu2024rdt,
  title={{RDT-1B}: A Diffusion Foundation Model for Bimanual Manipulation},
  author={Liu, Songming and Wu, Lingxuan and Li, Bangguo and Tan, Hengkai and Chen, Huayu and Wang, Zhengyi and Xu, Ke and Su, Hang and Zhu, Jun},
  booktitle={International Conference on Learning Representations},
  year={2025}
}

@inproceedings{lipman2023flowmatching,
  title={Flow Matching for Generative Modeling},
  author={Lipman, Yaron and Chen, Ricky TQ and Ben-Hamu, Heli and Nickel, Maximilian and Le, Matthew},
  booktitle={International Conference on Learning Representations},
  year={2023}
}

@inproceedings{wu2023gr1,
  title={Unleashing Large-Scale Video Generative Pre-training for Visual Robot Manipulation},
  author={Wu, Hongtao and Jing, Ya and Cheang, Chilam and Chen, Guangzeng and Xu, Jiafeng and Li, Xinghang and Liu, Minghuan and Li, Hang and Kong, Tao},
  booktitle={International Conference on Learning Representations},
  year={2024}
}

@inproceedings{li2024vision,
  title={Vision-Language Foundation Models as Effective Robot Imitators},
  author={Li, Xinghang and Liu, Minghuan and Zhang, Hanbo and Yu, Cunjun and Xu, Jie and Wu, Hongtao and Cheang, Chilam and Jing, Ya and Zhang, Weinan and Liu, Huaping and Li, Hang and Kong, Tao},
  booktitle={International Conference on Learning Representations},
  year={2024}
}

@article{bu2024towards,
  title={Towards synergistic, generalized, and efficient dual-system for robotic manipulation},
  author={Bu, Qingwen and Li, Hongyang and Chen, Li and Cai, Jisong and Zeng, Jia and Cui, Heming and Yao, Maoqing and Qiao, Yu},
  journal={arXiv preprint arXiv:2410.08001},
  year={2024}
}

@article{reuss2025flower,
  title={{FLOWER}: Democratizing Generalist Robot Policies with Efficient Vision-Language-Action Flow Policies},
  author={Reuss, Moritz and Zhou, Hongyi and R{\"u}hle, Marcel and Ya{\u{g}}murlu, {\"O}mer Erdin{\c{c}} and Otto, Fabian and Lioutikov, Rudolf},
  journal={arXiv preprint arXiv:2509.04996},
  year={2025}
}

@inproceedings{tian2025predictive,
  title={Predictive Inverse Dynamics Models Are Scalable Learners for Robotic Manipulation},
  author={Tian, Yang and Yang, Sizhe and Zeng, Jia and Wang, Ping and Lin, Dahua and Dong, Hao and Pang, Jiangmiao},
  booktitle={International Conference on Learning Representations},
  year={2025}
}

@article{li2025gr,
  title={{GR-MG}: Leveraging Partially-Annotated Data via Multi-Modal Goal-Conditioned Policy},
  author={Li, Peiyan and Wu, Hongtao and Huang, Yan and Cheang, Chilam and Wang, Liang and Kong, Tao},
  journal={IEEE Robotics and Automation Letters},
  volume={10},
  number={2},
  pages={1912--1919},
  year={2025},
  publisher={IEEE}
}

@inproceedings{du2023learning,
  title={Learning Universal Policies via Text-Guided Video Generation},
  author={Du, Yilun and Yang, Sherry and Dai, Bo and Dai, Hanjun and Nachum, Ofir and Tenenbaum, Josh and Schuurmans, Dale and Abbeel, Pieter},
  booktitle={Advances in Neural Information Processing Systems},
  volume={36},
  pages={9156--9172},
  year={2023}
}

@inproceedings{hu2024video,
  title={Video Prediction Policy: A Generalist Robot Policy with Predictive Visual Representations},
  author={Hu, Yucheng and Guo, Yanjiang and Wang, Pengchao and Chen, Xiaoyu and Wang, Yen-Jen and Zhang, Jianke and Sreenath, Koushil and Lu, Chaochao and Chen, Jianyu},
  booktitle={Proceedings of the 42nd International Conference on Machine Learning},
  pages={24328--24346},
  year={2025},
  volume={267},
  series={Proceedings of Machine Learning Research},
  publisher={PMLR},
  url={https://proceedings.mlr.press/v267/hu25g.html}
}

@misc{unifiedwm2025,
      title={Unified World Models: Coupling Video and Action Diffusion for Pretraining on Large Robotic Datasets}, 
      author={Chuning Zhu and Raymond Yu and Siyuan Feng and Benjamin Burchfiel and Paarth Shah and Abhishek Gupta},
      year={2025},
      eprint={2504.02792},
      archivePrefix={arXiv},
      primaryClass={cs.RO},
      url={https://arxiv.org/abs/2504.02792}, 
}

@article{ye2026dreamzero,
  title={World Action Models are Zero-Shot Policies},
  author={Ye, Seonghyeon and Ge, Yunhao and Zheng, Kaiyuan and Gao, Shenyuan and Yu, Sihyun and Kurian, George and Indupuru, Suneel and Tan, You Liang and Zhu, Chuning and Xiang, Jiannan and others},
  journal={arXiv preprint arXiv:2602.15922},
  year={2026}
}

@article{fastwam2026,
  title={{Fast-WAM}: Do World Action Models Need Test-Time Future Imagination?},
  author={Yuan, Tianyuan and Dong, Zibin and Liu, Yicheng and Zhao, Hang},
  journal={arXiv preprint arXiv:2603.16666},
  year={2026}
}

@misc{onetokenperframe2026,
      title={One Token Per Frame: Reconsidering Visual Bandwidth in World Models for VLA Policy}, 
      author={Zuojin Tang and Shengchao Yuan and Xiaoxin Bai and Zhiyuan Jing and De Ma and Gang Pan and Bin Liu},
      year={2026},
      eprint={2605.07931},
      archivePrefix={arXiv},
      primaryClass={cs.CV},
      url={https://arxiv.org/abs/2605.07931}, 
}

@article{lawam2026,
  title={{LaWAM}: Latent World Action Models for Efficient Dynamics-Aware Robot Policies},
  author={Chen, Jialei and Wang, Kai and Chen, Kang and Chen, Shuaihang and Gao, Feng and Tang, Wenhao and Li, Zhiyuan and Liu, Weilin and Yao, Zhuyu and Li, Boxun and others},
  journal={arXiv preprint arXiv:2606.15768},
  year={2026}
}

@article{repwam2026,
  title={RepWAM: World Action Modeling with Representation Visual-Action Tokenizers},
  author={Wang, Junke and Zhang, Qihang and Yang, Shuai and Luo, Yiming and Shen, Yujun and Wu, Zuxuan and Jiang, Yu-Gang and Xu, Yinghao},
  journal={arXiv preprint arXiv:2606.13674},
  year={2026}
}

@article{reconstruction_semantics2026,
  title={Reconstruction or Semantics? What Makes a Latent Space Useful for Robotic World Models},
  author={{Nilaksh} and Jha, Saurav and Zholus, Artem and Chandar, Sarath},
  journal={arXiv preprint arXiv:2605.06388},
  year={2026}
}

@inproceedings{lastzero2026,
  title={{LaST$_0$}: Latent Spatio-Temporal Chain-of-Thought for Robotic Vision-Language-Action Model},
  author={Liu, Zhuoyang and Liu, Jiaming and Chen, Hao and Yu, Jiale and Guo, Ziyu and Hou, Chengkai and Gu, Chenyang and Mi, Xiangju and Zhang, Renrui and Wu, Kun and others},
  booktitle={Forty-third International Conference on Machine Learning},
  year={2026}
}

@article{lda1b2026,
  title={{LDA-1B}: Scaling Latent Dynamics Action Model via Universal Embodied Data Ingestion},
  author={Lyu, Jiangran and Liu, Kai and Zhang, Xuheng and Liao, Haoran and Feng, Yusen and Zhu, Wenxuan and Shen, Tingrui and Chen, Jiayi and Zhang, Jiazhao and Dong, Yifei and others},
  journal={arXiv preprint arXiv:2602.12215},
  year={2026}
}

@article{dreamvla2025,
  title={{DreamVLA}: A Vision-Language-Action Model Dreamed with Comprehensive World Knowledge},
  author={Zhang, Wenyao and Liu, Hongsi and Qi, Zekun and Wang, Yunnan and Yu, Xinqiang and Zhang, Jiazhao and Dong, Runpei and He, Jiawei and Wang, He and Zhang, Zhizheng and others},
  journal={Advances in Neural Information Processing Systems},
  volume={38},
  pages={24195--24228},
  year={2025}
}

@misc{lecun2022ami,
  title={A Path Towards Autonomous Machine Intelligence},
  author={LeCun, Yann},
  year={2022},
  note={Version 0.9.2},
  url={https://openreview.net/pdf?id=BZ5a1r-kVsf}
}

@misc{assran2023ijepa,
      title={Self-Supervised Learning from Images with a Joint-Embedding Predictive Architecture}, 
      author={Mahmoud Assran and Quentin Duval and Ishan Misra and Piotr Bojanowski and Pascal Vincent and Michael Rabbat and Yann LeCun and Nicolas Ballas},
      year={2023},
      eprint={2301.08243},
      archivePrefix={arXiv},
      primaryClass={cs.CV},
      url={https://arxiv.org/abs/2301.08243}, 
}

@article{bardes2024vjepa,
  title={Revisiting Feature Prediction for Learning Visual Representations from Video},
  author={Bardes, Adrien and Garrido, Quentin and Ponce, Jean and Chen, Xinlei and Rabbat, Michael and LeCun, Yann and Assran, Mahmoud and Ballas, Nicolas},
  journal={arXiv preprint arXiv:2404.08471},
  year={2024}
}

@article{vjepa2_2025,
  title={{V-JEPA 2}: Self-Supervised Video Models Enable Understanding, Prediction and Planning},
  author={Assran, Mido and others},
  journal={arXiv preprint arXiv:2506.09985},
  year={2025}
}

@article{vjepa21_2026,
  title={{V-JEPA 2.1}: Unlocking Dense Features in Video Self-Supervised Learning},
  author={Mur-Labadia, Lorenzo and Muckley, Matthew and Bar, Amir and Assran, Mido and Sinha, Koustuv and Rabbat, Mike and LeCun, Yann and Ballas, Nicolas and Bardes, Adrien},
  journal={arXiv preprint arXiv:2603.14482},
  year={2026}
}

@misc{ghosh2018actionable,
      title={Learning Actionable Representations with Goal-Conditioned Policies}, 
      author={Dibya Ghosh and Abhishek Gupta and Sergey Levine},
      year={2019},
      eprint={1811.07819},
      archivePrefix={arXiv},
      primaryClass={cs.LG},
      url={https://arxiv.org/abs/1811.07819}, 
}

@article{vlajepa2026,
  title={{VLA-JEPA}: Enhancing Vision-Language-Action Model with Latent World Model},
  author={Sun, Jingwen and Zhang, Wenyao and Qi, Zekun and Ren, Shaojie and Liu, Zezhi and Zhu, Hanxin and Sun, Guangzhong and Jin, Xin and Chen, Zhibo},
  journal={arXiv preprint arXiv:2602.10098},
  year={2026}
}

@misc{disentangled_dynamics2026,
      title={Disentangled Robot Learning via Separate Forward and Inverse Dynamics Pretraining}, 
      author={Wenyao Zhang and Bozhou Zhang and Zekun Qi and Wenjun Zeng and Xin Jin and Li Zhang},
      year={2026},
      eprint={2604.16391},
      archivePrefix={arXiv},
      primaryClass={cs.RO},
      url={https://arxiv.org/abs/2604.16391}, 
}

@article{liu2023libero,
  title={{LIBERO}: Benchmarking Knowledge Transfer for Lifelong Robot Learning},
  author={Liu, Bo and Zhu, Yifeng and Gao, Chongkai and Feng, Yihao and Liu, Qiang and Zhu, Yuke and Stone, Peter},
  journal={Advances in Neural Information Processing Systems},
  volume={36},
  pages={44776--44791},
  year={2023}
}

@article{fei2025liberoplus,
  title={{LIBERO-Plus}: In-depth Robustness Analysis of Vision-Language-Action Models},
  author={Fei, Senyu and Wang, Siyin and Shi, Junhao and Dai, Zihao and Cai, Jikun and Qian, Pengfang and Ji, Li and He, Xinzhe and Zhang, Shiduo and Fei, Zhaoye and others},
  journal={arXiv preprint arXiv:2510.13626},
  year={2025}
}

@article{mees2021calvin,
  title={{CALVIN}: A Benchmark for Language-Conditioned Policy Learning for Long-Horizon Robot Manipulation Tasks},
  author={Mees, Oier and Hermann, Lukas and Rosete-Beas, Erick and Burgard, Wolfram},
  journal={IEEE Robotics and Automation Letters},
  volume={7},
  number={3},
  pages={7327--7334},
  year={2022},
  publisher={IEEE}
}
\bibliographystyle{references}

\appendix
\section{Appendix}
\label{sec:appendix}

\subsection{Detailed Original-LIBERO Results}
\label{app:libero-results}

Table~\ref{tab:libero-original-baselines} reports suite-level results on the original LIBERO benchmark.
The main paper combines the original-benchmark overall score with the zero-shot LIBERO-Plus perturbation results for a compact robustness comparison.

\begin{table*}[!htbp]
\caption{Original LIBERO baseline comparison. All numbers are success rates in percent. P.T. indicates additional embodied policy/world pretraining before target-suite training; generic visual-language or video backbone initialization alone is not counted.}
\label{tab:libero-original-baselines}
\centering
\small
\setlength{\tabcolsep}{2.4pt}
\renewcommand{\arraystretch}{1.08}
\begin{tabular}{lcc@{\hspace{3pt}}!{\vrule width 0.4pt}@{\hspace{3pt}}cccc@{\hspace{3pt}}!{\vrule width 0.4pt}@{\hspace{3pt}}c}
\gapthickhline
Method & P.T. & Size & Long & Spatial & Object & Goal & Overall \\
\gapthickhline
\multicolumn{8}{l}{\textit{VLA baselines}} \\
\gaphline
OpenVLA & \cmark & 7B & 53.7 & 84.7 & 88.4 & 79.2 & 76.5 \\
OpenVLA-OFT & \cmark & 7B & \third{94.5} & 97.6 & 98.4 & \third{97.9} & 97.1 \\
$\pi_0$ & \cmark & 3.3B & 85.2 & 96.8 & \third{98.8} & 95.8 & 94.1 \\
$\pi_{0.5}$ & \cmark & 3.3B & 92.4 & \third{98.8} & 98.2 & \second{98.0} & 96.9 \\
NORA & \cmark & 3B & 74.6 & 92.2 & 95.4 & 89.4 & 87.9 \\
WorldVLA & \xmark & 7B & 59.0 & 85.6 & 89.0 & 82.6 & 79.1 \\
UnifiedVLA & \cmark & 8.5B & 94.0 & 95.4 & \third{98.8} & 93.6 & 95.5 \\
RIPT-VLA & \cmark & 7B & 93.8 & \third{99.0} & 98.6 & \best{98.6} & \second{97.5} \\
\gapthickhline
\multicolumn{8}{l}{\textit{WAM baselines}} \\
\gaphline
Fast-WAM & \xmark & 6B & \second{95.2} & 98.2 & \best{100.0} & 97.0 & \best{97.6} \\
VLA-JEPA & \cmark & 3B & \best{95.8} & \best{99.6} & 96.2 & 97.2 & \third{97.2} \\
\gapthickhline
\multicolumn{8}{l}{\textit{Ours}} \\
\gaphline
\textbf{SLIM} & \xmark & \textbf{0.47B} & 94.4 & \second{99.4} & \second{99.4} & 96.8 & \second{97.5} \\
\gapthickhline
\end{tabular}%
\end{table*}

\subsection{Attention Analysis Details}
\label{app:interpretability-analysis}

The attention probe in Section~\ref{sec:experiments} is produced by our SLIM analysis pipeline.
It provides a diagnostic comparison between a Stage-1 pretrained policy and a direct policy trained without masked trajectory prediction.
The main paper reports benchmark numbers only from Figure~\ref{fig:ablation-lines}; the diagnostic pair is used to analyze internal behavior rather than to introduce an additional benchmark comparison.

\paragraph{Analysis pipeline.}
The paired pipeline captures matched diagnostic rollouts for the pretrained and direct policies and samples representative phases from each sequence.
For each frame, it extracts action-to-observation patch attention from layer 15, applies max aggregation over action tokens, and projects the resulting patch scores back to the image plane.
The observation row contains the unmodified frames, while dashed boxes identify task-relevant objects for visual reference.

\paragraph{Action-to-patch attention.}
The heatmaps visualize where the action stream retrieves observation-side information during policy inference.
They are intended as a qualitative diagnostic rather than a causal attribution.
The main-paper figure uses two matched examples to show how attention evolves from approaching a target to manipulating it.

\subsection{Model Parameterization}
\label{app:model-parameterization}

Table~\ref{tab:slim-parameter-breakdown} reports the parameter breakdown of the SLIM configuration used in our LIBERO experiments.
The DINOv2-B/14 visual encoder contains 86.58M parameters, while the action model contains 385.56M parameters.
The latter is dominated by the 16-layer MoT interaction trunk; its remaining 7.60M parameters comprise the action and state interfaces, learned token embeddings, conditioning projections, and prediction heads.
The model size reported in the main paper, 0.47B, corresponds to 472.14M trainable policy parameters.
The frozen T5-small language encoder is excluded from the reported 0.47B trainable-policy parameter count.

\begin{table}[!htbp]
\caption{Parameter breakdown of SLIM. Shares are computed relative to the 472.14M trainable parameters and the component rows are non-overlapping. The frozen EMA target encoder is instantiated only during Stage 1 and is excluded from the reported model size.}
\label{tab:slim-parameter-breakdown}
\centering
\small
\setlength{\tabcolsep}{5pt}
\renewcommand{\arraystretch}{1.08}
\begin{tabular}{lrr}
\gapthickhline
Component & Parameters & Trainable share \\
\gapthickhline
DINOv2-B/14 visual encoder & 86.58M & 18.34\% \\
MoT interaction trunk & 377.96M & 80.05\% \\
Action/state interfaces and prediction heads & 7.60M & 1.61\% \\
\gaphline
\textbf{Trainable policy total} & \textbf{472.14M} & \textbf{100.00\%} \\
\gapthickhline
Frozen EMA visual target encoder & 86.58M & -- \\
Stage-1 instantiated total & 558.72M & -- \\
\gapthickhline
\end{tabular}
\end{table}

For optimizer grouping, 471.58M trainable parameters are subject to weight decay, while 0.56M bias and normalization parameters use zero weight decay.
The frozen EMA encoder provides latent prediction targets in Stage 1, receives no gradients, and is not part of the active Stage-2 inference path.

\subsection{Implementation Details}
\label{app:implementation-details}

\paragraph{Inputs and model configuration.}
SLIM processes two RGB views---a workspace view and a wrist view---resized to $224\times224$.
The DINOv2-B/14 visual encoder is fine-tuned jointly with the policy, whereas the T5-small language encoder is frozen and its language embeddings are cached.
Actions are normalized using per-dataset 1st- and 99th-percentile statistics.
Both simulation benchmarks use 7-dimensional actions; the action horizon is 8 for LIBERO and 12 for CALVIN.
We use four flow-sampling steps at inference.

\paragraph{Optimization.}
We train the simulation policies in BF16 on eight NVIDIA H100 80GB GPUs with gradient clipping at 1.0 and no gradient accumulation.
We use AdamW with $\beta_1=0.9$, $\beta_2=0.95$, $\epsilon=10^{-8}$, and weight decay $0.01$; bias and normalization parameters receive no weight decay.
The action model and visual encoder use learning rates of $10^{-4}$ and $10^{-5}$, respectively, while the remaining trainable parameters use a base learning rate of $2.5\times10^{-5}$.
Learning rates follow cosine decay to a minimum of $10^{-6}$.
Stage 1 is trained for 3 epochs with a 2,000-step warmup, $\lambda_{\mathrm{IDM}}=0.125$, $\lambda_{\mathrm{FDM}}=1$, and EMA momentum $0.999$.
Stage 2 is initialized from the Stage-1 checkpoint and trained for 40 epochs using only the flow-matching policy objective; the EMA target encoder is not retained in Stage 2.
The real-world policy uses the same configuration, except that Stage 2 is trained for 20 epochs.

\paragraph{Benchmark-specific settings.}
For LIBERO, Stage 1 mixes LIBERO-90 with the four target suites, whereas Stage 2 uses only LIBERO-10, LIBERO-Object, LIBERO-Spatial, and LIBERO-Goal.
The global batch size is 128, and Stage 2 uses a 5,000-step warmup.
We evaluate 50 rollouts per task (2,000 rollouts over the four suites), then evaluate the same checkpoint without adaptation on all 10,030 LIBERO-Plus perturbation cases.
For CALVIN, both stages use the language-annotated ABC training split with a global batch size of 256 and a 2,000-step warmup.
Evaluation follows the standard 1,000 five-instruction chains in the held-out environment D.

\subsection{Policy Inference Benchmark Details}
\label{app:inference-benchmark}

\paragraph{Setup.}
All measurements use one exclusive NVIDIA H100 80GB HBM3 at batch size 1 with two deterministic $224\times224$ RGB inputs, BF16 precision, and the fixed instruction ``do something.''
The benchmark measures model inference only; model loading, simulator stepping, and RPC are excluded.
Each latency run uses 20 warmups followed by 200 synchronized calls, repeated in three fresh processes for 600 pooled measurements.
Peak VRAM is measured in a separate process as incremental device memory after model, input, and language-cache setup; model loading, language encoding, and profiler runs are excluded.

\paragraph{Model configurations and FLOPs.}
SLIM predicts an 8-step action chunk with 4 sampling steps, $\pi_{0.5}$ predicts a 10-step chunk with 10 sampling steps, and Fast-WAM predicts a 32-step chunk with 10 sampling steps.
SLIM and Fast-WAM use precomputed language embeddings, while $\pi_{0.5}$ follows its native PyTorch policy path.
All three models run in PyTorch eager mode, and FLOPs are measured per action chunk using \texttt{torch.profiler} with math SDPA.
Because native horizons differ, the reported values correspond to one replanning call or action chunk rather than one executed action.

\subsection{Real-World Evaluation Details}
\label{app:real-world-details}

\begin{figure}[t]
\centering
\begin{minipage}[t]{0.44\textwidth}
\vspace{0pt}
\textbf{Task-specific OOD settings.}
Figure~\ref{fig:real-world-ood-details} shows the nominal setup and all three visual perturbations for each of the five real-world tasks.
The background condition replaces the nominal tabletop texture, the lighting condition projects colored light patterns over the workspace, and the distractor condition introduces task-irrelevant objects while preserving the task goal.
The same task-condition configurations are used for all evaluated policies.
\end{minipage}
\hfill
\begin{minipage}[t]{0.52\textwidth}
\vspace{0pt}
\centering
\includegraphics[width=\linewidth]{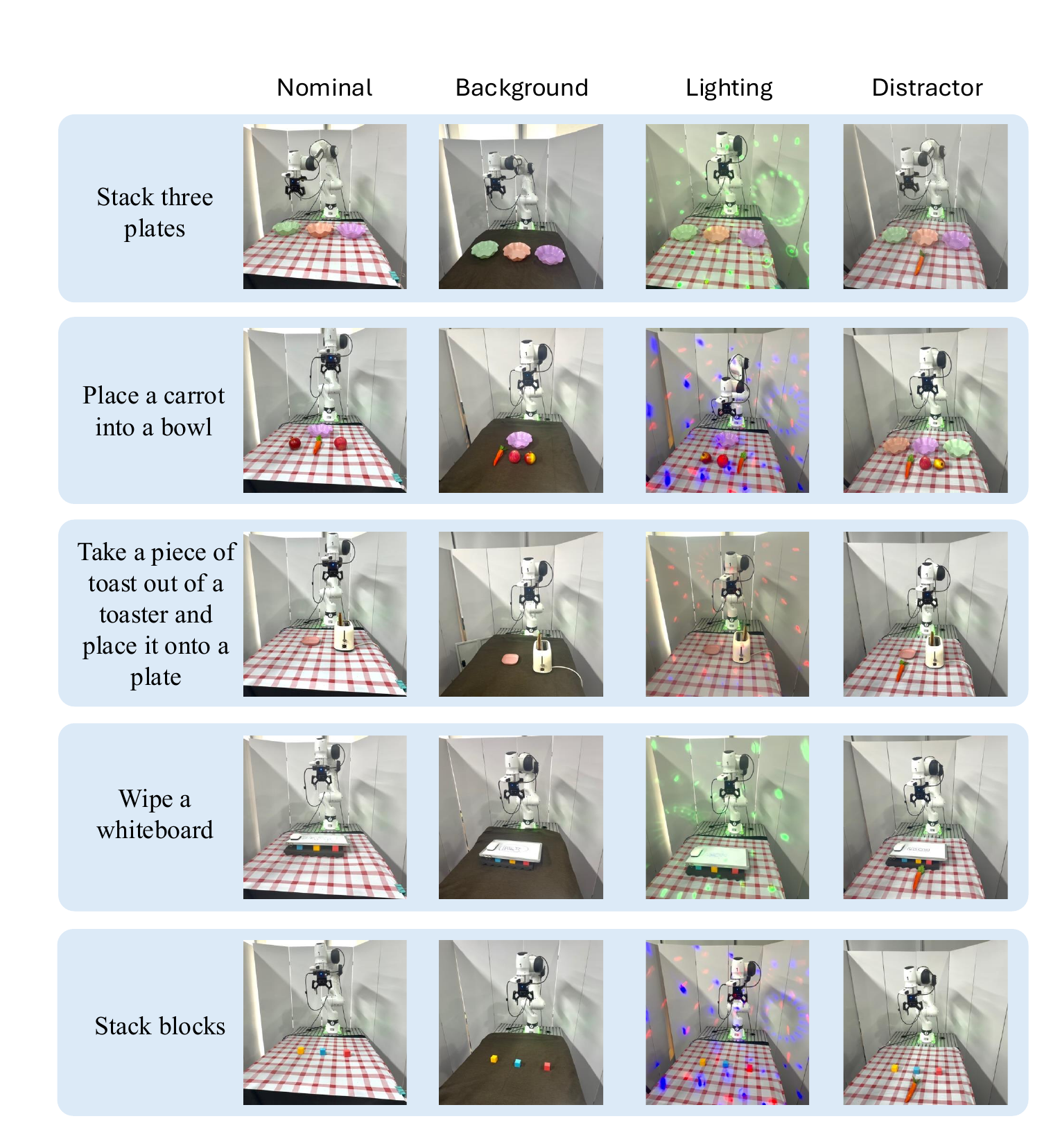}
\caption{Task-specific real-world evaluation settings. Columns show the nominal setup and the background, lighting, and distractor perturbations.}
\label{fig:real-world-ood-details}
\end{minipage}
\end{figure}

\begin{figure}[t]
\centering
\begin{minipage}[t]{0.44\textwidth}
\vspace{0pt}
\textbf{Progress-score definitions.}
Each trial receives a score in $\{0, 0.5, 1\}$ according to the furthest verified task milestone: 0 indicates that neither milestone is reached, 0.5 indicates partial completion, and 1 indicates full completion.
For stacking three plates, forming a stack of two plates scores 0.5, while stacking all three scores 1.
For placing a carrot into a bowl, grasping the carrot scores 0.5, while placing it into the bowl scores 1.
For the toast task, grasping the toast scores 0.5, while placing it onto the plate scores 1.
For wiping the whiteboard, grasping the eraser scores 0.5, while completing the wipe scores 1.
For stacking blocks, completing the first stacking action scores 0.5, while completing the second and forming the full stack scores 1.
Figure~\ref{fig:real-world-task-progression} illustrates the corresponding initial, partial-progress, and full-completion stages.
The main-paper progress score is computed as 100 times the mean trial score.
\end{minipage}
\hfill
\begin{minipage}[t]{0.52\textwidth}
\vspace{0pt}
\centering
\includegraphics[width=\linewidth]{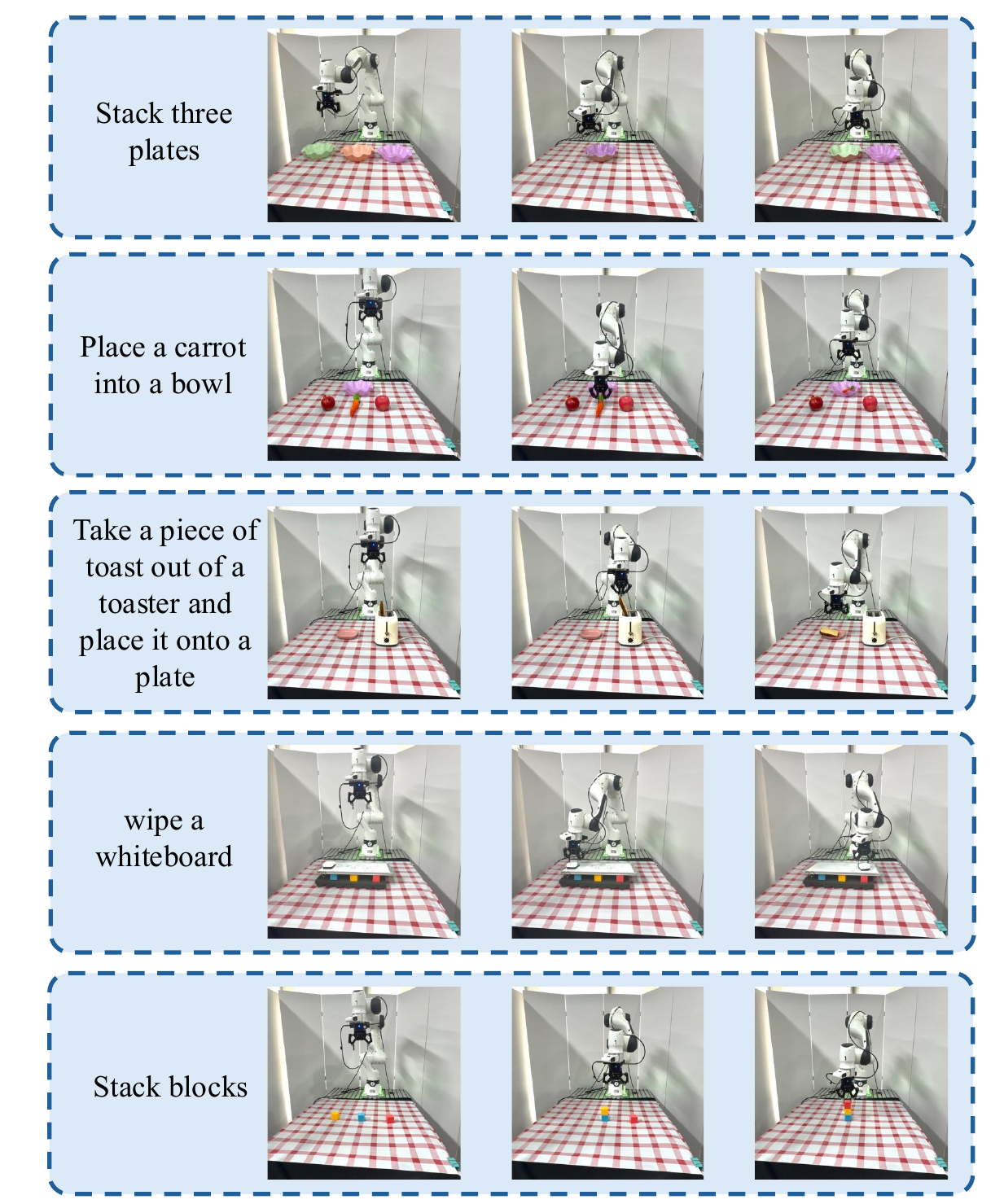}
\caption{Progress-score milestones for the five real-world tasks. Each row shows the initial state, partial completion (score 0.5), and full completion (score 1).}
\label{fig:real-world-task-progression}
\end{minipage}
\end{figure}

\begin{figure*}[t]
\centering
\includegraphics[width=0.96\textwidth]{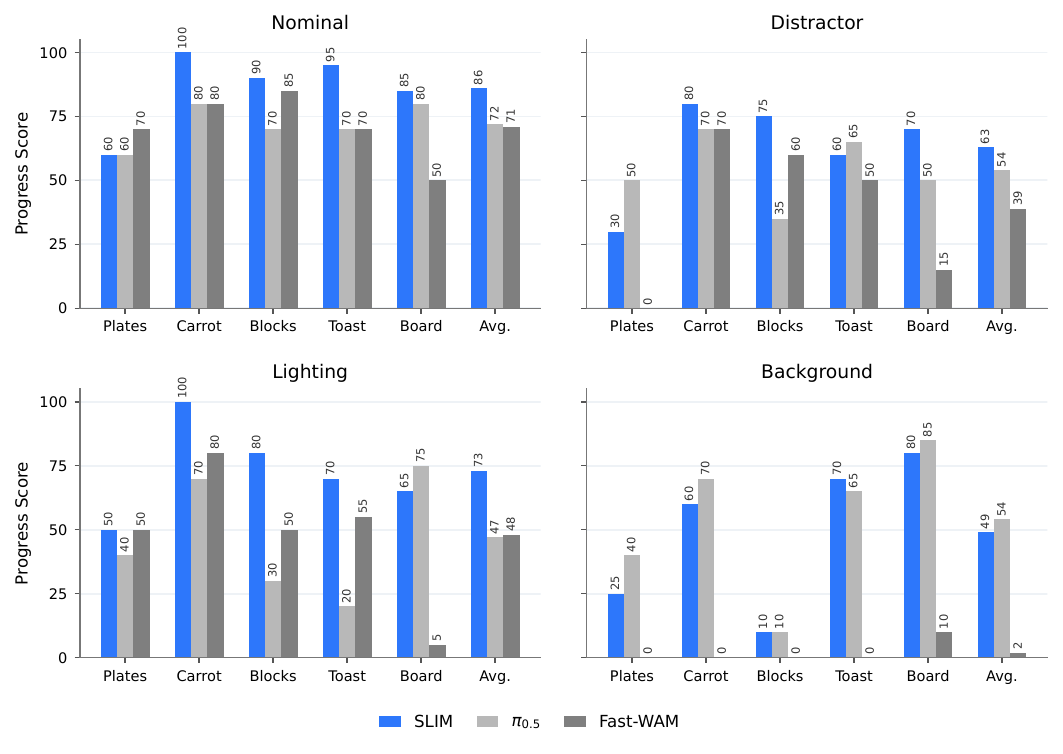}
\caption{Task-wise real-world progress under the nominal setting and the distractor, lighting, and background perturbations. Each panel reports all five tasks and their average over 10 trials per task.}
\label{fig:real-world-progress-detailed}
\end{figure*}

\paragraph{Per-task results.}
Figure~\ref{fig:real-world-progress-detailed} reports the complete task-wise progress scores under the nominal setting and each OOD perturbation.
The rightmost group in each panel is the average across the five tasks and corresponds to the setting-level values summarized in the main paper.

\end{document}